\pdfoutput=1
\documentclass[pdflatex,sn-basic]{sn-jnl}

\usepackage{graphicx}
\usepackage{amsmath,amssymb}
\usepackage{booktabs}
\usepackage{tabularx}
\usepackage{xcolor}
\usepackage{textcomp}
\usepackage{tikz}
\usetikzlibrary{arrows.meta,positioning}
\usepackage{subcaption}
\usepackage{CJKutf8}

\newcommand{\NPagesAnnotated}{1{,}082}

\newcommand{\NPagesReal}{658}

\newcommand{\NRefs}{787}
\newcommand{\NRefsYear}{639}

\newcommand{\NMockReal}{3}

\newcommand{\NCandPagesCorpus}{1{,}005}
\newcommand{\NExtraLow}{89}
\newcommand{\NCandComicsCorpus}{684}

\newcommand{\NDupComics}{4}

\newcommand{\CicPre}{4}\newcommand{\CicPreLo}{2}\newcommand{\CicPreHi}{8}\newcommand{\CicPreN}{182}
\newcommand{\CicWar}{11}\newcommand{\CicWarLo}{7}\newcommand{\CicWarHi}{17}\newcommand{\CicWarN}{137}
\newcommand{\CicPost}{3}\newcommand{\CicPostLo}{1}\newcommand{\CicPostHi}{5}\newcommand{\CicPostN}{304}
\newcommand{\FaceSit}{57}\newcommand{\FaceSitLo}{48}\newcommand{\FaceSitHi}{66}\newcommand{\FaceSitN}{116}
\newcommand{\FaceHist}{3}\newcommand{\FaceHistLo}{1}\newcommand{\FaceHistHi}{7}\newcommand{\FaceHistN}{143}

\newcommand{\CovHorrorFifties}{26}

\newcommand{\CovHorrorWar}{2}

\newcommand{\CovSuperheroFifties}{6}

\newcommand{\CovSuperheroWar}{42}

\newcommand{\DetFOne}{0.97}

\newcommand{\GcdPrecision}{72}

\newcommand{\HeurFOne}{0.66}

\newcommand{\MPanelsEighties}{5.3}

\newcommand{\MPanelsNoughties}{5.0}
\newcommand{\MSplashEighties}{11}

\newcommand{\MSplashNoughties}{15}
\newcommand{\NComics}{3{,}942}

\newcommand{\NDated}{3{,}034}

\newcommand{\NGcdMatched}{2{,}215}

\newcommand{\NGrammarComics}{3{,}929}
\newcommand{\NGrammarPages}{195{,}795}

\newcommand{\NSampledPages}{27{,}018}
\newcommand{\NTrendComics}{2{,}924}

\newcommand{\NetSamePub}{32}
\newcommand{\NetSamePubExp}{11}

\newcommand{\PanelsLvlPre}{7.28}
\newcommand{\PanelsLvlWar}{6.02}

\newcommand{\PanelsWar}{-1.26}
\newcommand{\PanelsWarHi}{-1.14}
\newcommand{\PanelsWarLo}{-1.38}
\newcommand{\PctDated}{77}
\newcommand{\ProbeAcc}{70}
\newcommand{\ProbeMaj}{27}

\newcommand{\SplashWar}{4.0}
\newcommand{\SplashWarHi}{4.5}
\newcommand{\SplashWarLo}{3.5}

\newcommand{\StyleArtists}{351}
\newcommand{\StyleComics}{1{,}165}

\newcommand{\StylePages}{4{,}497}
\newcommand{\WordsFifties}{75}
\newcommand{\WordsFiftiesHi}{85}
\newcommand{\WordsFiftiesLo}{65}
\newcommand{\WordsLvlFifties}{264}

\newcommand{\WordsLvlWar}{189}

\newcommand{\WordsWar}{21}
\newcommand{\WordsWarHi}{28}
\newcommand{\WordsWarLo}{14}
\newcommand{\ZsAcc}{46}

\newcommand{\VPagesSampled}{1{,}132}
\newcommand{\VPages}{1{,}055}
\newcommand{\VBalloons}{10{,}282}

\newcommand{\VWordShare}{15.7}

\newcommand{\VWpbF}{14.2}
\newcommand{\VWpbM}{13.7}

\newcommand{\VRelN}{269}
\newcommand{\VRelAgree}{99.6}
\newcommand{\VRelKappa}{0.99}

\newcommand{\VWordPre}{11}

\newcommand{\VWordWar}{10}

\newcommand{\VWordPost}{20}

\newcommand{\VWordFifties}{18}

\newcommand{\VWordNoLead}{10}
\newcommand{\VNNoLead}{913}
\newcommand{\VWordLead}{51}
\newcommand{\VNLead}{142}

\newcommand{\VRoleLoveInterestF}{12.9}
\newcommand{\VRoleLoveInterestM}{0.9}
\newcommand{\VRoleAuthorityF}{0.7}
\newcommand{\VRoleAuthorityM}{7.0}
\newcommand{\VToMaleF}{65}

\newcommand{\VToFemaleF}{12}

\newcommand{\VAdjWordWarPre}{+2.4}
\newcommand{\VAdjWordWarPreLo}{\textminus{}2.3}
\newcommand{\VAdjWordWarPreHi}{6.6}

\newcommand{\VAdjWordPostPre}{+13.3}
\newcommand{\VAdjWordPostPreLo}{8.5}
\newcommand{\VAdjWordPostPreHi}{18.5}

\newcommand{\VAdjGapPostPre}{+5.0}
\newcommand{\VAdjGapPostPreLo}{0.6}
\newcommand{\VAdjGapPostPreHi}{8.0}

\newcommand{\MVols}{87}
\newcommand{\MPages}{7{,}235}
\newcommand{\MChars}{2{,}684}

\newcommand{\MRelN}{167}
\newcommand{\MRelAgree}{96.4}
\newcommand{\MRelKappa}{0.93}

\newcommand{\MBrPages}{112}
\newcommand{\MBrBal}{1{,}154}
\newcommand{\MBrAgree}{94.2}
\newcommand{\MBrKappa}{0.88}
\newcommand{\MBrUS}{39.5}
\newcommand{\MBrLink}{41.6}

\newcommand{\MWordAll}{40.3}
\newcommand{\MWordAllLo}{34.9}
\newcommand{\MWordAllHi}{45.6}

\newcommand{\MGapAll}{\textminus{}0.3}

\newcommand{\MWordUS}{15.7}
\newcommand{\MWordUSLo}{14.2}
\newcommand{\MWordUSHi}{17.3}

\newcommand{\MGapUS}{1.0}

\newcommand{\MWordShonen}{34.7}

\newcommand{\MWordSeinen}{39.2}

\newcommand{\MWordShojo}{49.4}

\newcommand{\MGapShojo}{4.0}
\newcommand{\MGapShojoLo}{1.2}
\newcommand{\MGapShojoHi}{6.6}
\newcommand{\MWordJosei}{41.7}

\newcommand{\XEvalPages}{102}
\newcommand{\XEvalComics}{98}
\newcommand{\XChance}{0.32}
\newcommand{\XClip}{0.54}
\newcommand{\XRawBM}{0.34}
\newcommand{\XShufBM}{0.34}
\newcommand{\XDenseExp}{0.47}

\newcommand{\XPackage}{501}
\newcommand{\XOffset}{261}
\newcommand{\XJudgePages}{45}
\newcommand{\XJudgeKappa}{0.58}

\newcommand{\XClipRaw}{+0.20}
\newcommand{\XClipRawLo}{0.14}
\newcommand{\XClipRawHi}{0.25}
\newcommand{\XClipDenseExp}{+0.06}
\newcommand{\XClipDenseExpLo}{\textminus{}0.02}
\newcommand{\XClipDenseExpHi}{0.15}

\newcommand{\XDenseShuf}{+0.14}
\newcommand{\XDenseShufLo}{0.09}
\newcommand{\XDenseShufHi}{0.20}

\newcommand{\GStories}{215{,}961}

\newcommand{\GNoPencil}{22}

\newcommand{\GPrinted}{10}
\newcommand{\GUnsure}{8}

\newcommand{\DateValN}{146}
\newcommand{\DateValExact}{139}
\newcommand{\DateValOne}{145}

\newcommand{\StrictUniqueN}{4{,}264}
\newcommand{\StrictComicHit}{7.6}
\newcommand{\StrictComicNull}{2.2}

\newcommand{\StrictUniqueHit}{2.4}
\newcommand{\StrictUniqueNull}{1.5}
\newcommand{\StrictUniqueLift}{1.6}
\newcommand{\StrictUniqueP}{<0.001}

\newcommand{\HNamedChars}{215}
\newcommand{\HNamedFemale}{91}
\newcommand{\HPageN}{454}
\newcommand{\HPageAgree}{92.7}
\newcommand{\HCropN}{202}

\newcommand{\HCropAgree}{95.5}
\newcommand{\HCropUnclear}{13}
\newcommand{\HWomenAsMen}{17}
\newcommand{\HMenAsWomen}{16}
\newcommand{\HShareHuman}{32.6}
\newcommand{\HShareCoded}{30.4}
\newcommand{\HPageElig}{477}
\newcommand{\HPageSpeakers}{92}
\newcommand{\HPageLo}{85}
\newcommand{\HPageHi}{98}
\newcommand{\HGapLo}{\textminus{}14.6}
\newcommand{\HGapHi}{+6.8}
\newcommand{\MRelAll}{86}
\newcommand{\MRelTotal}{200}

\begin{document}

\title[Studying comics at scale with AI]{Studying Comics at Scale with Artificial Intelligence: The PAGE Framework and Three Studies of American Comics and Japanese Manga}

\author*[1]{\fnm{Emilio} \sur{Ferrara}}\email{emiliofe@usc.edu}

\affil*[1]{\orgdiv{Thomas Lord Department of Computer Science}, \orgname{University of Southern California}, \orgaddress{\city{Los Angeles}, \state{CA}, \country{USA}}}

\abstract{Comics are one of the great visual records of the twentieth century: hundreds of
thousands of pages in which a mass readership saw its leaders, its wars and its heroes. Most of this record has never been studied as a whole; comics scholarship proceeds one book at
a time. We argue that artificial intelligence, including generative models, can change this to the benefit of the medium itself. We propose PAGE, a framework that makes artificial intelligence an instrument for comics research. It parses every scan into
panels, text and a bibliographic record; annotates each page with a vision-language model that
follows a written codebook, as a human coder would; grounds samples of these readings in human annotation; and explains change across an archive, with estimates controlled for publisher and genre. We demonstrate the framework on \NGrammarComics{} American comic books
(1938--1956) and \MVols{} volumes of Japanese manga, in three studies, each resting on a different capability. The first pairs text search with visual reading to recover how a mass
medium portrayed political power. The second measures the page's visual form across an industry and recovers the hand of individual artists. The third follows who speaks across two comics traditions, checked against human annotation. Beyond
research, the same instruments can make archives searchable by what is drawn, open comics to blind and low-vision readers, and help preserve fragile collections and credit uncredited artists. We outline a research agenda in which artificial intelligence strengthens
the study, preservation and reach of the medium.}

\keywords{comics, manga, vision-language models, cultural analytics, computational social science, generative AI}

\maketitle

\section{Introduction}\label{sec:intro}

In June 1941, six months before Pearl Harbor, a reader of \emph{Pep Comics} \#16 watched a
president hand a young hero his commission in the United States Navy. The page never prints the president's name; the reader was expected to recognize Franklin Roosevelt by his pince-nez and his profile. Hundreds of thousands of such pages survive. A single public collection of golden-age American comic books, the COMICS
corpus \citep{iyyer2017amazing}, holds \NGrammarComics{} books and \NGrammarPages{} pages,
and the Grand Comics Database \citep{gcd2026,monaco2016grand} indexes \GStories{} comic stories in American issues dated 1938 to 1956. Together these pages record how a mass readership saw its leaders, its wars and its heroes,
and how a new visual language was invented on cheap newsprint. The questions this record
raises concern all of it: how a popular medium reflected and shaped the society that read it,
how its visual language formed and spread across publishers, artists and countries, and whose
stories it told and whose voices it carried. No scholar can read enough pages to answer them.

Comics scholarship has answered such questions by careful reading of selected books and,
more recently, by hand-coded corpora of a few dozen titles
\citep{cohn2013visual,dunst2018empirical,pederson2016changing,cohn2017picture}. Computer
vision has meanwhile learned to find panels, balloons and characters on a page
\citep{augereau2018survey,nguyen2018digital,sachdeva2024manga}, and vision-language models
can read a page image and answer questions about it in plain language
\citep{vivoli2024comix,baek2026mangavqa}. These two lines of work have rarely met, because detection models are evaluated on benchmarks and comics scholars seldom have instruments they can trust at the scale of a whole archive. We ask whether artificial intelligence can serve as a trustworthy instrument for studying the comics medium, and whether the indexes, transcripts and attribution evidence it produces can also help the medium's readers, archives and artists.

We answer with PAGE, a framework that organizes the work into four steps, and with three studies that apply it. \emph{Parse} turns each scan into structured objects: panels, text boxes, reading
order, and a bibliographic record with date, publisher and credited artists. \emph{Annotate}
has a vision-language model read each page under a written codebook, the same document a
human coder would follow, and records what the page shows in the categories the codebook defines, such as how a figure
is drawn, what role a character plays and who speaks each line. \emph{Ground} checks samples of those annotations against evidence
the model did not produce: a second coder with different information, human annotation where
it exists, quantities read by an independent method, and close reading of the pages a finding
rests on. \emph{Explain} turns annotations into trends with publisher, genre and sampling held fixed, with intervals that resample comics or volumes, and reads those trends against the history of the medium to answer the study's question. Built this way, artificial intelligence answers questions about the medium that require reading every page of an archive, and it produces indexes searchable by what is drawn, transcripts that a screen reader can voice and evidence for crediting artists whose names were never printed, all of which can serve the medium's readers, archives and artists.

Our contributions are four. (i) We define the PAGE framework and catalog the techniques from
computer vision, language technology and generative AI that serve each step, together with
uses for readers, archives and artists: discovery, attribution, access, preservation and
teaching. (ii) We demonstrate it on \NGrammarComics{} American comic books and \MVols{} volumes of
Japanese manga in three studies, each an instance of one of these questions and each built on a
different capability of the framework. To ask how the medium reflected its society, we pair a search of the lettering with a reading of the drawing and recover how golden-age comics portrayed the President of the United States. To ask how its visual language formed, we measure the visual form of every page, trace how the American page changed between 1938 and 1956 and match pages to their credited pencillers. To ask whose voices it carried, we attribute dialogue to speakers and compare who speaks in American comics and in manga. (iii) We show how to ground
model annotation in practice, with a second coder working from different information, gender
stated by human annotators as a reference for the model's speaker coding, and checks of dates
and panel counts by independent methods. (iv) We set out a research agenda of six lines of work, from discovery by what is drawn to shared human reference sets, in which these instruments serve scholarship and the medium's readers, archives and artists.

\section{Related work}\label{sec:related}

\paragraph{Comics as an object of empirical study.} Comics theory describes a page as a sequence of panels whose meaning arises in the gaps between them \citep{mccloud1993understanding}, and cultural studies read comics as a record of material and everyday culture \citep{jenkins2020comics}. Cohn's account of comics as a visual language, with its own vocabulary and grammar and with distinct American and Japanese dialects, gave the field testable hypotheses \citep{cohn2013visual,cohn2019your,cohn2021who}, and empirical comics research tests them on annotated corpora \citep{dunst2018empirical,laubrock2019computational}. Hand-coded studies
compare panels in American comics and manga \citep{cohn2011different,cohn2012framing}, trace
page layouts and the balance of text and image in American superhero comics across eight
decades \citep{pederson2016changing,cohn2017picture}, and count how much of the dialogue male and female characters speak in manga written for boys and for girls \citep{unserschutz2015what}. Annotated
collections include eBDtheque \citep{guerin2013ebdtheque}, the Graphic Narrative Corpus
\citep{dunst2017graphic}, COMICS \citep{iyyer2017amazing} and Manga109
\citep{matsui2016sketch,aizawa2020building}, with speaker links in Manga109Dialog
\citep{li2024manga109dialog} and onomatopoeia in COO \citep{baek2022coo}. An early example of
large-scale cultural analysis of manga images is Manovich's visualization of a million
fan-translated pages \citep{manovich2011understanding}, and distant viewing has applied
computer vision to comic strips such as \emph{Peanuts} \citep{wigard2023understanding}.

\paragraph{Computer vision and language technology for comics.} A survey covers panel and balloon detection, text recognition, character recognition and reader interaction \citep{augereau2018survey}, and indexing pipelines combine these detectors to make comics searchable \citep{nguyen2018digital}. Recent models detect panels, text and
characters jointly and associate each balloon with its speaker
\citep{sachdeva2024manga,sachdeva2024tails,li2024zeroshot}, translate manga using the image
as context \citep{hinami2021towards}, and are evaluated on benchmarks for detection, reading
order, speaker identification and narrative closure
\citep{vivoli2024comix,vivoli2025comicspap,ikuta2025mangaub,baek2026mangavqa}. On the task of choosing the panel that completes a strip, general multimodal models perform near chance \citep{vivoli2025comicspap}, which is one reason we check the model's annotations against independent evidence before we count them.

\paragraph{Models as measurement instruments for culture.} Quantitative cultural history was
popularized by counts of words in digitized books \citep{michel2011quantitative} and grew into
cultural analytics and distant viewing of images
\citep{manovich2020cultural,arnold2019distant,lee2020newspaper}. It now spans narrative and
visual media: the emotional arcs of more than a thousand works of fiction reduce to six basic shapes
\citep{reagan2016emotional}, explainable image embeddings trace the aesthetic complexity of
125,000 artworks and recognize their authorship and style \citep{karjus2023compression}, and machine learning follows artistic influence across more than 11,000 Japanese Ukiyo-e prints
\citep{honna2026disruptive}. Computational folkloristics linked large archives of traditional narrative to scholarly categories
in searchable graphs \citep{abello2012computational}. Computational studies of
fiction and film measure who appears and who speaks, finding women under-represented among
characters \citep{underwood2018transformation,kraicer2019social} and portrayed differently in their dialogue \citep{ramakrishna2017linguistic} and in the power and agency of their actions \citep{sap2017connotation}; an AI agent, with human-in-the-loop validation, has annotated how female main characters are portrayed in 13,869 novels
\citep{kratzke2026female}. Vision-language models have also labeled images at scale, with the labels checked by hand \citep{chen2025synthetic}. Large language models
now code text at or above the accuracy of crowd workers \citep{gilardi2023chatgpt,tornberg2023chatgpt4},
can apply a qualitative codebook at scale \citep{dunivin2025scaling}, perform best as an aid to human annotators \citep{ziems2024can}, carry demographic and cultural biases \citep{ferrara2023chatgpt}, and can
change their answers with the surface form of a prompt \citep{ye2026robustness}. We therefore calibrate a model that reads comics before we use its codes, in the Ground step of our framework.

\paragraph{Generative AI and the people who make comics.} Generative image models raise
questions of authorship, labor and consent for artists \citep{epstein2023art}; they have been
trained on artists' work without permission and used to imitate living artists' styles \citep{jiang2023ai}. Zhao and colleagues have built two protections that artists can apply to their own work. Glaze adds small perturbations that cloak an artwork's style from mimicry by text-to-image models \citep{shan2023glaze}, and Nightshade turns images scraped for training without consent into poison for the models trained on them, as a defense for content owners \citep{shan2024nightshade}. The same group
finds that expert artists and automated detectors each tell human art from AI-generated images
well but err in different ways, and argues for combining them \citep{ha2024organic}. Synthetic media also threaten the provenance of digital archives
\citep{ferrara2024genai,ferrara2026paradox}. Generative tools for comics include
reference-based colorization \citep{furusawa2017comicolorization} and models that generate
manga panels with consistent characters \citep{wu2025diffsensei}. We take these concerns as
design constraints: every use of generative AI that we propose serves scholarship, access or preservation, and nothing in this work should be read as an endorsement or
enablement of artwork generation with AI.

\section{The PAGE framework}\label{sec:framework}

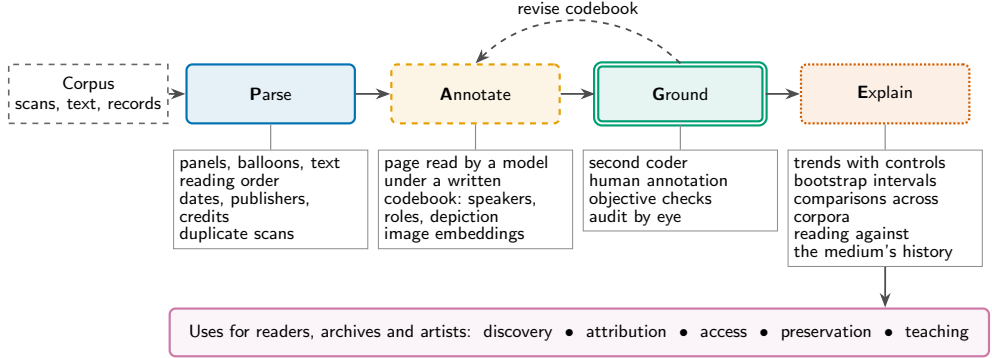
\begin{figure}[!htbp]
\centering
\resizebox{\linewidth}{!}{\begin{tikzpicture}[
  font=\sffamily\small,
  stage/.style={rectangle, rounded corners=3pt, draw=#1, line width=1.1pt, fill=#1!10,
                minimum width=3.0cm, minimum height=1.05cm, align=center},
  obj/.style={rectangle, draw=black!45, fill=white, text width=3.25cm, align=left,
              font=\sffamily\scriptsize, inner sep=3pt},
  arr/.style={-{Stealth[length=2.6mm]}, line width=0.9pt, draw=black!70},
  ]
\definecolor{oiBlue}{HTML}{0072B2}\definecolor{oiOrange}{HTML}{E69F00}
\definecolor{oiGreen}{HTML}{009E73}\definecolor{oiVerm}{HTML}{D55E00}
\definecolor{oiPurple}{HTML}{CC79A7}

\node[rectangle, draw=black!60, dashed, line width=0.8pt, minimum width=2.3cm, minimum height=1.05cm,
      align=center] (corpus) at (-3.25,0) {Corpus\\[-1pt]\scriptsize scans, text, records};
\node[stage=oiBlue] (P) at (0,0) {\textbf{P}arse};
\node[stage=oiOrange, dashed] (A) at (3.65,0) {\textbf{A}nnotate};
\node[stage=oiGreen, double, double distance=1pt] (G) at (7.3,0) {\textbf{G}round};
\node[stage=oiVerm, densely dotted] (E) at (10.95,0) {\textbf{E}xplain};

\draw[arr] (corpus) -- (P);
\draw[arr] (P) -- (A);
\draw[arr] (A) -- (G);
\draw[arr] (G) -- (E);
\draw[arr, dashed] (G.north) to[out=135, in=45] node[above, font=\sffamily\scriptsize] {revise codebook} (A.north);

\node[obj, below=0.45cm of P] (oP) {panels, balloons, text\\reading order\\dates, publishers,\\credits\\duplicate scans};
\node[obj, below=0.45cm of A] (oA) {page read by a model\\under a written\\codebook: speakers,\\roles, depiction\\image embeddings};
\node[obj, below=0.45cm of G] (oG) {second coder\\human annotation\\objective checks\\audit by eye};
\node[obj, below=0.45cm of E] (oE) {trends with controls\\bootstrap intervals\\comparisons across\\corpora\\reading against\\the medium's history};
\foreach \s/\o in {P/oP, A/oA, G/oG, E/oE} \draw[black!45, line width=0.6pt] (\s) -- (\o);

\node[rectangle, rounded corners=3pt, draw=oiPurple, line width=1.1pt, fill=oiPurple!8,
      minimum width=14.6cm, minimum height=0.85cm, align=center] (R) at (5.5,-4.25)
      {Uses for readers, archives and artists: \ discovery \ \textbullet\ \ attribution \ \textbullet\ \ access \ \textbullet\ \ preservation \ \textbullet\ \ teaching};
\draw[arr] (oE.south) -- (oE.south |- R.north);
\end{tikzpicture}}
\caption{The PAGE framework. Parse turns scans, text and bibliographic records into structured objects; Annotate has a vision-language model read each page under a written codebook and adds image embeddings; Ground checks samples of the annotations against independent evidence, and disagreements found there lead to revisions of the codebook; Explain estimates trends with publisher, genre and sampling held fixed and reads them against the history of the medium to answer the study's question. The same objects can serve readers, archives and
artists through discovery, attribution, access, preservation and teaching.}
\label{fig:framework}
\end{figure}

PAGE uses a vision-language model as empirical research uses any measurement instrument
(Figure~\ref{fig:framework}). A written codebook fixes what the model records before any page
is read, and we check a sample of the model's codes against human annotation and other independent
evidence before using them to answer a question about the medium.

\paragraph{Parse.} In the Parse step we detect the panels and text boxes of each page, transcribe the lettering, order the boxes for reading, and attach the page to a bibliographic record. In COMICS, panel and text boxes and an OCR
transcription come with the release \citep{iyyer2017amazing}; for manga we train a Faster
R-CNN panel detector \citep{ren2017faster} on the hand-drawn frames of Manga109-s, which
reaches box F1 of \DetFOne{} on held-out books, against \HeurFOne{} for simple gutter rules.
Without a bibliographic record, a page cannot be dated or assigned to a publisher, and public comics corpora often lack these records. The COMICS release carries no dates, publishers or
credits, so we read each book's cover with an on-device text recognizer and match the series
and issue printed in the indicia to the Grand Comics Database (GCD). We also detect the same printed issue scanned twice, so that we count it once.

\paragraph{Annotate.} In the Annotate step we record what each page shows, in categories fixed before the reading.
A vision-language model reads the page image, together with any text the parser extracted, under a written codebook that defines each category, its admissible values and its rules for hard cases. We use Claude Opus 5.5 (Anthropic) for every
codebook annotation in this paper, run as independent sessions that each read a fixed set of
pages. For the manga characters, we discarded a first pass in which some sessions were given more images than they could display, and recoded those characters with one task per session. In the speaker study we shuffle pages across tasks and withhold their years, so that any drift in the model's coding during a run cannot align with publication year; the model read the presidents pages in batches in corpus order. We also compute CLIP ViT-B/32 image embeddings \citep{radford2021learning}, which place each embedded page in a space where nearness means visual similarity and support retrieval, artist attribution and genre classification, as image representations recognize authorship and style in fine art \citep{karjus2023compression}.

\paragraph{Ground.} In the Ground step we test whether the model's annotations are correct against four kinds of evidence that the model did not produce. A second coder recodes a sample from different information, for example the drawings without the character's name, and we report agreement with Cohen's $\kappa$ \citep{cohen1960coefficient}. Where human annotation exists, we score the coding against it directly. Manga109 annotators named many minor characters by their gender ( We test the parser with quantities read by an independent method, comparing dates read from the cover by text recognition with dates read from the indicia by the vision model, and detected panels with hand-drawn frames. We check any case a finding rests on, such as a president identified by his face alone, by close reading of the page. We also revise the codebook in response to grounding evidence; the first version of our speaker codebook recorded gender, age and species in one field, and a first coding of 32 pages led us to separate them before the main run.

\paragraph{Explain.} In the Explain step we estimate change in the annotations and read it against what is known of the medium's history.
We make causal claims only where the design supports them. First, every corpus is a selection, because the publishers and genres that survive vary by year, and a trend in the raw means can therefore be a change in the collection. We fit each
measure on year with publisher-group and genre fixed effects and report adjusted yearly means.
Second, pages of one comic are not independent, so intervals come from a bootstrap that
resamples comics or volumes \citep{efron1993introduction}. We report contrasts between named windows (for example 1940--41 against 1944--45), so that each claim names the years it compares. We then read the estimates against the history of the medium, such as the war, the campaign against comics that led to the Comics Code, and the readerships for which manga are published.

\paragraph{Uses for readers, archives and artists.} Table~\ref{tab:catalog} lists the techniques that serve each PAGE step and each use for readers, archives and artists, and marks which ones this paper tests. \emph{Discovery}: archives can become searchable by what is drawn, as Newspaper Navigator made searchable the photographs and illustrations it extracted from sixteen million newspaper pages \citep{lee2020newspaper}.
\emph{Attribution}: of the \GStories{} American comic stories of 1938--1956 in the GCD,
\GNoPencil\% name no penciller, \GPrinted\% of the stories with a structured penciller credit are flagged as credited in the comic itself, and \GUnsure\% of credited stories carry a
credit marked uncertain; most penciller credits for the period therefore rest on later
research. Image embeddings that match pages to their penciller (Section~\ref{sec:page}) can add a visual signal to that research. \emph{Access}: reading order, speaker links
and transcripts, the products of Parse and Annotate, are what a screen reader or an audio
description needs for a reader who cannot see the page \citep{sachdeva2024manga}, and
image-aware translation opens a book to readers of another language
\citep{hinami2021towards}. \emph{Preservation}: checksummed, indexed scans keep fragile
newsprint retrievable, and restored or colorized access copies
\citep{furusawa2017comicolorization} can be kept apart from the original scans with their
provenance recorded. \emph{Teaching}: pages located by what they show become primary sources
for the history of a period, as the presidents study illustrates.

\begin{table}[!htbp]
\caption{AI techniques for studying the comics medium, by PAGE step and by use for readers,
archives and artists. The last column gives the use in this paper; a dash marks techniques we
catalog but did not use.}
\label{tab:catalog}
\footnotesize
\begin{tabularx}{\linewidth}{@{}l>{\raggedright\arraybackslash}X>{\raggedright\arraybackslash}X>{\raggedright\arraybackslash}X@{}}
\toprule
Step & Task & Technique and example tools & Use in this paper \\
\midrule
Parse & Panels and text boxes & Object detection (Faster R-CNN \citep{ren2017faster}); joint panel, text and character detection \citep{sachdeva2024manga} & Panel detector trained on Manga109-s; COMICS release boxes \\
 & Lettering & OCR; on-device text recognition; transcription models \citep{sachdeva2024manga} & COMICS OCR; cover dates \\
 & Bibliographic records & Text recognition on covers and indicia; record linkage to the GCD \citep{gcd2026} & Dates for \PctDated\% of comics; GCD publisher, genre, penciller \\
 & Duplicate scans & Perceptual hashing with close inspection & Duplicates among the Study~1 comics \\
Annotate & Page content under a codebook & Vision-language model reading the page image and extracted text & Presidents, speakers, roles, character gender \\
 & Visual similarity & Image embeddings (CLIP \citep{radford2021learning}) & Retrieval, artist matching, cover genre \\
 & Speaker attribution & Speaker detection and character re-identification \citep{li2024manga109dialog,li2024zeroshot,sachdeva2024tails} & Human speaker links (Manga109Dialog) \\
 & Text retrieval & Lexical (BM25) and dense sentence embeddings, query expansion & Retrieval comparison (Section~\ref{sec:grounding}) \\
Ground & Agreement & Second coder from different information; Cohen's $\kappa$ \citep{cohen1960coefficient} & Speaker and character coding \\
 & Accuracy & Human annotation as reference; independent methods for dates and counts & Gender-named manga characters; dates; panel frames \\
Explain & Trends with controls; historical reading & Fixed-effects regression; comic- and volume-level bootstrap \citep{efron1993introduction}; interpretation against the history of the medium & All three studies \\
\midrule
Use & Discovery & Search by what is drawn \citep{lee2020newspaper} & -- \\
 & Attribution & Nearest-neighbor artist matching on embeddings & Tested on credited pages \\
 & Access & Reading order, speaker links and transcripts for screen readers; image-aware translation \citep{sachdeva2024manga,hinami2021towards} & -- \\
 & Preservation & Colorization of access copies \citep{furusawa2017comicolorization}; checksummed archives & Lossless, checksummed mirror of all scans \\ & Teaching & Retrieval of primary-source pages by subject & Presidents study \\
 & Protection & Style cloaking and data poisoning against unauthorized training \citep{shan2023glaze,shan2024nightshade}; detection of AI-generated art \citep{ha2024organic} & -- \\
\bottomrule
\end{tabularx}
\end{table}

\paragraph{Principles of practice.} Six rules govern the studies below. (1) Every number
traces to annotations and every annotation to a page key, so any claim can be checked on the
scan. (2) The codebook is written before the main reading and revised only in response to
grounding evidence. (3) Grounding uses, where possible, information the model did not have or
a reference the model did not produce; Table~\ref{tab:grounding} reports which checks meet
this standard. (4) A printed issue scanned twice is counted once wherever it has been
detected. (5) We read Manga109-s pages under the dataset's research license. (6) We document model use as a method, naming the model and keeping its codebooks with the data.

\section{Data}\label{sec:data}

\paragraph{COMICS.} The COMICS release provides scans, panel and text boxes, and an OCR
transcription with each text box labeled as dialogue or narration, for golden-age American
comic books from the Digital Comic Museum \citep{iyyer2017amazing}. Of its \NComics{} books,
\NGrammarComics{} (\NGrammarPages{} pages) carry OCR and panel boxes. We mirror all scans
losslessly, recompressing each JPEG to JPEG~XL so that every page decodes to a byte-identical
copy, and index every page by checksum. Both the scans and the Grand Comics Database index used below
are the volunteer work of fans and collectors, part of the participatory culture that has long
organized comics fandom \citep{jenkins1992textual,jenkins2006convergence}.

\paragraph{Reconstructed metadata.} The release carries no bibliographic metadata. We read each
comic's cover and first pages with an on-device text recognizer and take the year printed
beside a month name or a copyright notice. Compared with the year that the vision model read
from the printed cover and indicia of the comics in Study~1, this rule agrees exactly for
\DateValExact{} of \DateValN{} comics and within one year for \DateValOne. Combining this rule with the scan releases' own text files and the vision model's reading of the indicia, we assign a year to \NDated{} comics (\PctDated\% of the release). We match each comic to the GCD (Library of Congress snapshot of 15 April 2026) by the series and issue printed in the indicia, and by year, and link \NGcdMatched{} comics; on comics whose title and issue we know independently, the matched issue is exact in \GcdPrecision\% of cases. The GCD supplies story genres and, story by story,
the credited penciller.

\paragraph{Manga.} Manga109-s provides \MVols{} Japanese volumes published from the 1970s to the
2010s, with hand-drawn boxes for every panel, text balloon, face and body
\citep{matsui2016sketch,aizawa2020building}; the dataset's catalog gives each title's decade,
readership (sh\=onen for boys, seinen for men, sh\=ojo for girls, josei for women) and genre.
Manga109Dialog links each balloon to the character who speaks it \citep{li2024manga109dialog}.

Appendix~\ref{app:data} documents two properties of the COMICS release that affect any study joining its parts, a page-numbering offset in its panel-annotation subset and duplicate scans of the same printed issue.

\section{Grounding the instrument before use}\label{sec:grounding}

Before we use the model's codes in the three studies, we check samples of them against human
annotation and evidence read by independent methods; Table~\ref{tab:grounding} lists every check
by study.

First, the model's gender coding matches the gender stated by human annotators. Manga109 annotators gave \HNamedChars{} characters names that state their gender (\HNamedFemale{} of them women), most of them minor characters such as ``woman 1'' or ``mother''. Because the character coding used in Study~3, which codes each manga character once from its crops and its name, saw the names, we recode their crop sheets with the name withheld for this test, and the model gives the annotators' gender for \HCropAgree\% of the \HCropN{} characters it classifies as a woman or a man (\HCropUnclear{} are coded unclear).
We also run the page-level procedure that Study~3 applies to American comics (Section~\ref{sec:voice}) on 112 random manga pages and 112 pages chosen to contain such balloons. The model sees each page and its balloons without character names, and it gives the annotators' gender for the speaker of \HPageAgree\% of the \HPageN{} balloons spoken by these characters that it codes as a woman or a man (95\% CI \HPageLo--\HPageHi\%, resampling the \HPageSpeakers{} speakers; \HPageElig{} balloons were eligible). The model's errors go in both directions (\HWomenAsMen{} women coded as men, \HMenAsWomen{} men coded as women), and on these balloons it gives women \HShareCoded\% of the text against \HShareHuman\% under the human reference, a difference whose interval includes zero (\HGapLo{} to \HGapHi{} points). Second, codings made under different information or codebook versions agree. In American comics, a first coding of 32 pages under the first codebook version and the
main coding under the revised version agree on the speaker's gender for \VRelAgree\% of the
\VRelN{} balloons that both coded as a woman or a man ($\kappa=\VRelKappa$); for manga
characters, a coder who sees the drawings without the name agrees with the main coder on
\MRelAgree\% of the \MRelN{} of \MRelTotal{} sampled characters that both coded as a woman or
a man ($\kappa=\MRelKappa$; \MRelAll\% exact agreement across all categories). Third, the
page-level procedure and the character coding agree on manga pages: on \MBrPages{} random
pages whose balloons come from the human text annotation, the speaker gender assigned on the
page matches the coded gender of the linked character for \MBrAgree\% of \MBrBal{} balloons
($\kappa=\MBrKappa$), and the two give women \MBrUS\% and \MBrLink\% of the text. 

\begin{table}[!htbp]
\caption{Grounding evidence by study. Human references are marked; the other checks compare
two readings made with different information or by different methods.}
\label{tab:grounding}
\footnotesize
\begin{tabularx}{\linewidth}{@{}>{\raggedright\arraybackslash}p{1.6cm}>{\raggedright\arraybackslash}X>{\raggedright\arraybackslash}p{2.5cm}@{}}
\toprule
Study & Check & Result \\
\midrule
Data & Cover-date OCR rule vs vision-model reading of indicia & \DateValExact/\DateValN{} exact \\
 & GCD issue match on known titles & \GcdPrecision\% exact \\
 & Manga panel detector vs hand-drawn frames (human) & F1 \DetFOne \\
Page & Cover genre classifier vs GCD genres (human), cross-validated & \ProbeAcc\% (majority \ProbeMaj\%) \\
 & Penciller matching vs GCD credits (human), other series, reprints excluded & \StrictUniqueHit\% vs \StrictUniqueNull\% \\
Speakers & Character gender, names withheld, vs gender-named characters (human) & \HCropAgree\%, $n=\HCropN$ \\
 & Page procedure vs gender-named speakers (human) & \HPageAgree\%, $n=\HPageN$ \\
 & Two coders, speaker gender (US) & $\kappa=\VRelKappa$, $n=\VRelN$ \\
 & Names withheld vs main coder, character gender (manga) & $\kappa=\MRelKappa$, $n=\MRelN$ \\
 & Page procedure vs character coding (manga) & \MBrAgree\%, $\kappa=\MBrKappa$ \\
\bottomrule
\end{tabularx}
\end{table}

\paragraph{Lessons from retrieval.} We also apply grounding to the data, in a comparison of text and image retrieval on the panel-annotation subset of COMICS.
Claude Opus 5.5 judged the relevance of pages to twelve scene queries (an airplane is drawn, characters fight, a concealed person is revealed) on \XEvalPages{} evaluation pages from \XEvalComics{} comics; one query had no relevant page. On a subset of \XJudgePages{} pages, OpenAI's \texttt{gpt-6-astra}
judged the same queries independently, and the two judges agree with $\kappa=\XJudgeKappa$. Lexical retrieval over raw OCR scores a macro average
precision of \XRawBM, no better than the same index with texts shuffled between pages
(\XShufBM) and close to chance (\XChance). The cause is a mismatch of vocabulary, because queries written in an analyst's words, such as  Expanding queries into the idiom of the comics and embedding text densely
raises text retrieval to \XDenseExp, above the shuffled lexical control by \XDenseShuf{}
(\XDenseShufLo{} to \XDenseShufHi). CLIP page retrieval reaches \XClip, above raw OCR by
\XClipRaw{} (\XClipRawLo{} to \XClipRawHi), and its advantage over dense retrieval with expanded queries is not distinguishable from zero (\XClipDenseExp, \XClipDenseExpLo{} to \XClipDenseExpHi). An
earlier run of the same comparison had joined images to text by the subset's page keys, which point to the wrong page for \XOffset{} of \XPackage{} images (Appendix~\ref{app:data}); the misalignment had overstated the advantage of images.

\section{Study 1: how a mass medium portrayed political power}\label{sec:presidents}

\paragraph{Goal and design.} In this study we pair a search of the lettering with a reading of the drawing, the first capability that PAGE brings to comics research, so that we can ask of an entire archive how a figure is depicted. We ask how American
comic books of 1938 to 1956 pictured the President of the United States, and whether the
picture changed with the Second World War. We search the OCR of every page for presidential
names and titles, which yields \NCandPagesCorpus{} candidate pages in \NCandComicsCorpus{}
comics. The vision model reads each candidate page image under a codebook and records who appears, whether he is drawn or only named, how his face is shown, his narrative role and the tone of the treatment. It rejects presidents of
companies, clubs and invented republics. Together with \NExtraLow{} sampled pages from
lower-precision search terms, and after removing duplicate scans, the model reads
\NPagesAnnotated{} pages, of which \NPagesReal{} carry a real presidential reference, for \NRefs{} references (\NRefsYear{} of them dated, a few before 1938). Dates come from the reading checked in Section~\ref{sec:grounding}; the portrayal codes have no second coder yet, a limitation we return to in Section~\ref{sec:discussion}.

\begin{figure}[!htbp]
\centering
\includegraphics[width=0.8\linewidth]{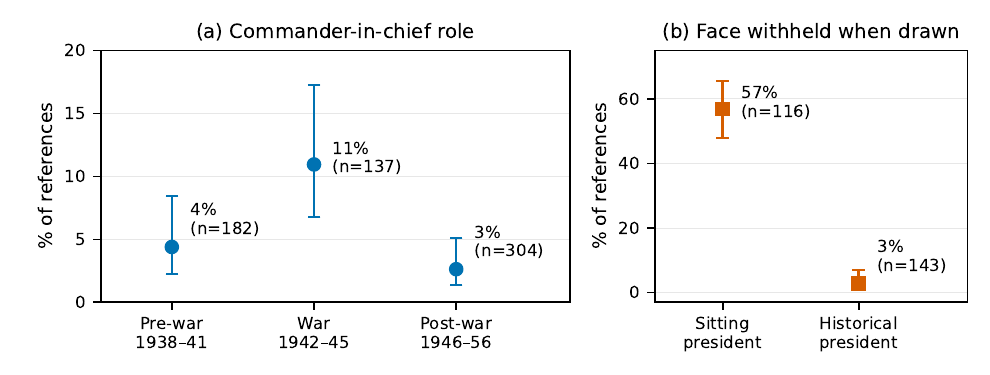}
\caption{The President in golden-age comics. Left: share of references in the
commander-in-chief role, by publication period. Right: among references that draw a
president, the share whose face is withheld, for sitting and historical presidents. Points are
proportions with Wilson 95\% intervals; $n$ is the number of references, with duplicate scans
of one printed issue counted once. Example pages are in Appendix~\ref{app:gallery}
(Figure~\ref{fig:plates}).}
\label{fig:presidents}
\end{figure}

\paragraph{Findings.} First, \emph{sitting presidents are drawn with the face withheld}. Among references that draw the sitting president, \FaceSit\% withhold his face
(95\% CI \FaceSitLo--\FaceSitHi\%, $n=\FaceSitN$): he appears as a silhouette, in shadow, from
the back of the head, or as a voice from behind a desk. Historical presidents are drawn this
way in \FaceHist\% of references (\FaceHistLo--\FaceHistHi\%, $n=\FaceHistN$); Washington,
Lincoln and Theodore Roosevelt appear as full portraits (Figure~\ref{fig:presidents}), and
part of this group are stamps, statues and wall portraits, whose faces are shown by
convention. Several pages show Roosevelt's face and never print his name (Figure~\ref{fig:plates}). Second, \emph{the commander-in-chief role peaks in wartime}.
The share of references that cast the president as commander-in-chief is \CicPre\% before
Pearl Harbor (\CicPreLo--\CicPreHi\%, $n=\CicPreN$), \CicWar\% in comics dated 1942--45
(\CicWarLo--\CicWarHi\%, $n=\CicWarN$) and \CicPost\% after the war (\CicPostLo--\CicPostHi\%,
$n=\CicPostN$); the wartime share is clearly above the postwar share, and its interval
overlaps slightly with the prewar one. Before the war the president is chiefly a national
symbol or the authority who dispatches the hero; after it, most references are passing
mentions, history features and advertisements for commemorative stamps. Third, \emph{portrayal
is deferential}. Of \NRefs{} references, \NMockReal{} mock a real president: a 1921 quip about
ex-President Taft in a humor magazine, which falls outside the period, a 1948 gag in which the
president hushes an aide to follow a baseball game, and a 1950 caricature. Every other mocking
or negative portrayal attaches to a stand-in, such as an impostor dressed as Washington in a
Nazi plot or a president hypnotized by a villain.

\paragraph{What PAGE enables.} Text search finds pages that name the president, and the model's reading of the page image records how he is portrayed. Text alone does not support the first finding, because a portrait and a
silhouette carry the same name in the lettering, and pages where Roosevelt is recognizable by
his face alone enter the candidate set only when another presidential term appears on them.
For this kind of question we therefore pair a search of the lettering with the model's reading of the drawing, and an image search for unnamed likenesses is the next step toward finding pages that name no president.

\section{Study 2: the visual form of the page across an industry}\label{sec:page}

\paragraph{Goal and design.} In this study we measure visual form at the scale of an industry. We describe every page by quantities such as panels, splash pages and words, and a fixed sample of pages by an image embedding, so that we can trace changes in the design of the medium, and the work of individual artists, across all surviving books. We ask how the design of the American
comic-book page changed between 1938 and 1956, and whether the change holds once we account
for which publishers and genres dominate each year. For every page of COMICS we compute, from the release's panel and
text boxes, the number of panels, whether a single panel covers at least 60\% of the panel area
(a splash), and words of lettering. For each measure we fit, across \NTrendComics{} dated
comics, a regression on year with publisher-group and genre fixed effects, and report adjusted
yearly means with comic-level bootstrap intervals. We compute CLIP image embeddings for a fixed sample of \NSampledPages{} pages (the cover and six interior pages per comic). Section~\ref{sec:grounding} checks the cover-genre classifier and the penciller labels against Grand Comics Database records.

\paragraph{Findings.} \emph{During the war, pages carried fewer and larger panels.} Panels per
page fall from \PanelsLvlPre{} in 1940--41 to \PanelsLvlWar{} in 1944--45, a change of
$\PanelsWar$ panels (95\% CI $\PanelsWarLo$ to $\PanelsWarHi$) with publisher and genre held
fixed, and splash pages rise by \SplashWar{} points (\SplashWarLo{} to \SplashWarHi).
\emph{After the war, words per page rose.} They rise modestly during the war (+\WordsWar, \WordsWarLo{} to \WordsWarHi) and then by \WordsFifties{} words (\WordsFiftiesLo{} to \WordsFiftiesHi) between 1944--45 and 1953--54, from \WordsLvlWar{} to \WordsLvlFifties, and the share of text boxes given to narration returns to its prewar level. Raw and adjusted curves nearly coincide (Figure~\ref{fig:trends}; typical pages of each period are in Appendix Figure~\ref{fig:typical}), and the contrasts keep their sign and
significance when we drop comics dated only through the GCD and when we restrict to comics of
known genre, so shifts in the publisher or genre mix of the surviving corpus account for little of either trend. Over eight decades of American superhero comics, hand-coded samples of 40 books show pages moving away from grid layouts and panels carrying fewer words \citep{pederson2016changing,cohn2017picture}; our whole-corpus estimates cover the first two decades of that span, in which words per page rose after the war.

\begin{figure}[!htbp]
\centering
\includegraphics[width=\linewidth]{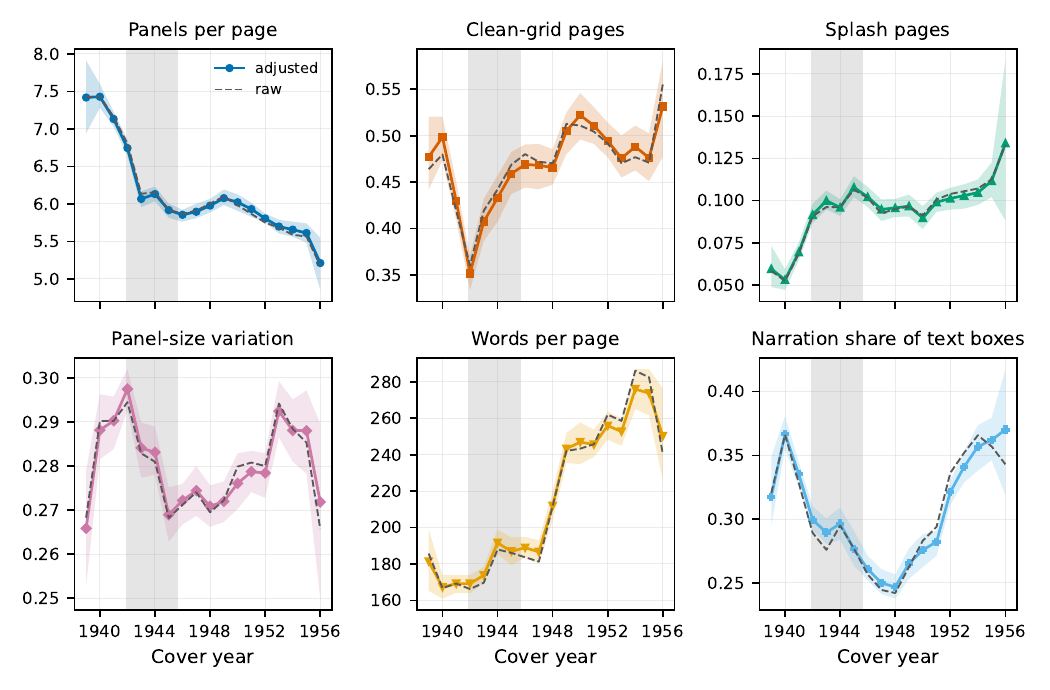}
\caption{Page grammar of American comic books by cover year, 1938--1956. Solid lines are
adjusted yearly means from comic-level regressions with publisher-group and genre fixed
effects, with comic-level bootstrap 95\% intervals; dashed lines are raw means. The shaded band
marks the United States at war. 1938 is pooled with 1939.}
\label{fig:trends}
\end{figure}

\begin{figure}[!htbp]
\centering
\includegraphics[width=\linewidth]{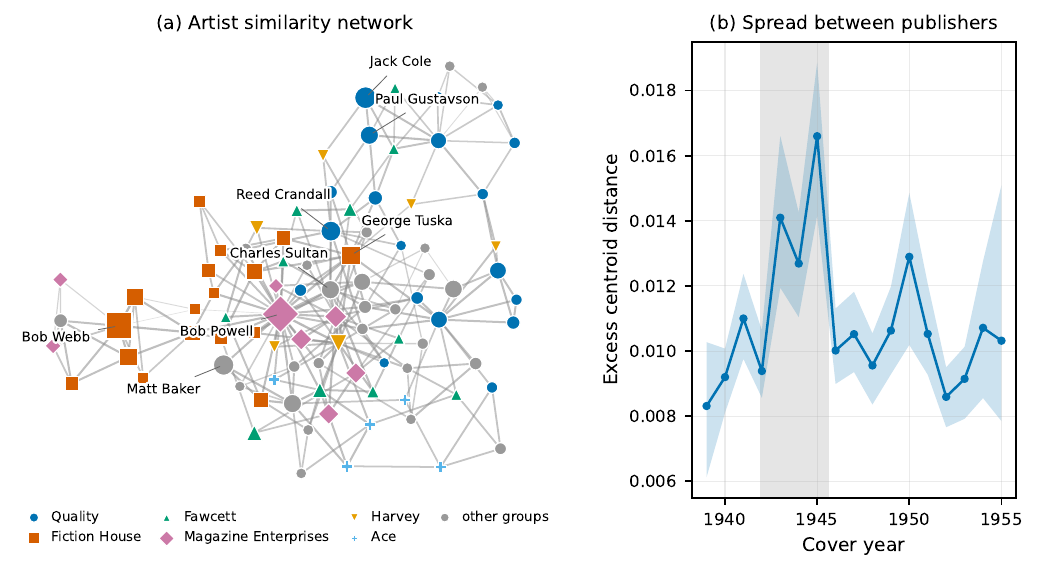}
\caption{(a) Artist similarity network: pencillers with at least 15 labeled pages, each joined
to its three nearest artists by mean CLIP embedding (Kamada--Kawai layout). Node size gives
the artist's labeled pages; marker shape and color give the main publisher group; the eight
artists with the most labeled pages are named. (b) Excess distance between publisher-group
centroids by year (observed minus label-shuffled), with 80\% subsampling intervals.}
\label{fig:style}
\end{figure}

\emph{Image embeddings match pages to their penciller.} We map sampled pages to stories, and stories to their GCD penciller, which labels \StylePages{} pages by \StyleArtists{} artists in \StyleComics{} comics. A page's nearest neighbor among pages of other comics is by the same
credited penciller \StrictComicHit\% of the time, against \StrictComicNull\% when labels are
shuffled within publisher and year. Much of this signal comes from other issues of the same series, which share characters and
logos, and from art printed more than once. When we exclude neighbors from the same series,
known duplicate scans, reprints of the same art (shared lettering or near-identical images),
and every page that nearly duplicates a page of another comic, such as a repeated
advertisement, \StrictUniqueN{} pages remain, and the rate is \StrictUniqueHit\% against
\StrictUniqueNull\%, a lift of \StrictUniqueLift{} over the shuffled baseline ($p\StrictUniqueP$; matched pairs in Appendix Figure~\ref{fig:pairs}). In the network of artists joined to their most similar peers,
\NetSamePub\% of links join artists of the same publisher group, against \NetSamePubExp\%
expected by chance, so pages also cluster by publisher group (Figure~\ref{fig:style}). \emph{Covers
record the turn from superheroes to horror.} A linear classifier on cover embeddings, trained
on GCD genres, names the genre of a held-out cover correctly \ProbeAcc\% of the time (majority
class \ProbeMaj\%, zero-shot text prompts \ZsAcc\%); applied to every dated cover, superhero
covers fall from \CovSuperheroWar\% in 1942--45 to \CovSuperheroFifties\% in 1950--56 while
horror covers rise from \CovHorrorWar\% to \CovHorrorFifties\%, the turn that culminated in the campaign against comics \citep{wertham1954seduction} and the Comics Code of 1954 \citep{nyberg1998seal,wright2001comic}. \emph{Later manga carry fewer panels per page.} In Manga109-s, panels per page fall from \MPanelsEighties{} in 1980s volumes to \MPanelsNoughties{} in 2000s volumes and splash pages rise from \MSplashEighties\% to \MSplashNoughties\%, the direction the American page took in the 1940s.

\paragraph{What PAGE enables.} Hand-coded studies of page layout over time rest on a few dozen books \citep{pederson2016changing,cohn2017picture}, and we extend them to every surviving page, with the composition of the corpus held fixed. The image embeddings also rank pages by their artist, so a corpus of mostly uncredited work can support research on attribution.

\section{Study 3: who speaks in two comics traditions}\label{sec:voice}

\paragraph{Goal and design.} In this study we attribute dialogue to speakers in two traditions and check the model's codes against human annotation, so that we can ask the same question about representation of American comics and of Japanese manga. We
ask which characters speak the dialogue, and whether the answer differs between the two
traditions. We treat each speech balloon as a unit of dialogue with a speaker, an addressee and a word count. For American comics we draw, from
each of up to 65 dated comics per year between 1938 and 1956, one interior story page with 3
to 25 dialogue balloons, for \VPagesSampled{} pages. The vision model reads each page with the
list of its dialogue balloons from the OCR and records, for every balloon, the speaker as the
character is drawn and presented in the story (gender, age, species), whom the speaker
addresses and the speaker's narrative role, together with the number of women and men drawn on
the page. The analysis covers \VPages{} story pages and \VBalloons{} speech balloons. For manga, Manga109Dialog identifies the speaker of each balloon, so the model codes each of the \MChars{} annotated characters once, from a sheet of crops of that character and the annotators' name for it. The comparison uses the \MPages{} manga pages that match the American
sampling frame: interior pages with 3 to 25 text boxes and at least one linked balloon.
Japanese is written without spaces, so manga shares are shares of text length in characters. Section~\ref{sec:grounding} shows that the speaker-gender coding gives the gender stated by human annotators for \HPageAgree\% of balloons.
We exclude speakers coded as groups or as unclear from all shares (example pages from both traditions in Appendix Figures~\ref{fig:usvoice} and~\ref{fig:mangapages}).

\begin{figure}[!htbp]
\centering
\includegraphics[width=\linewidth]{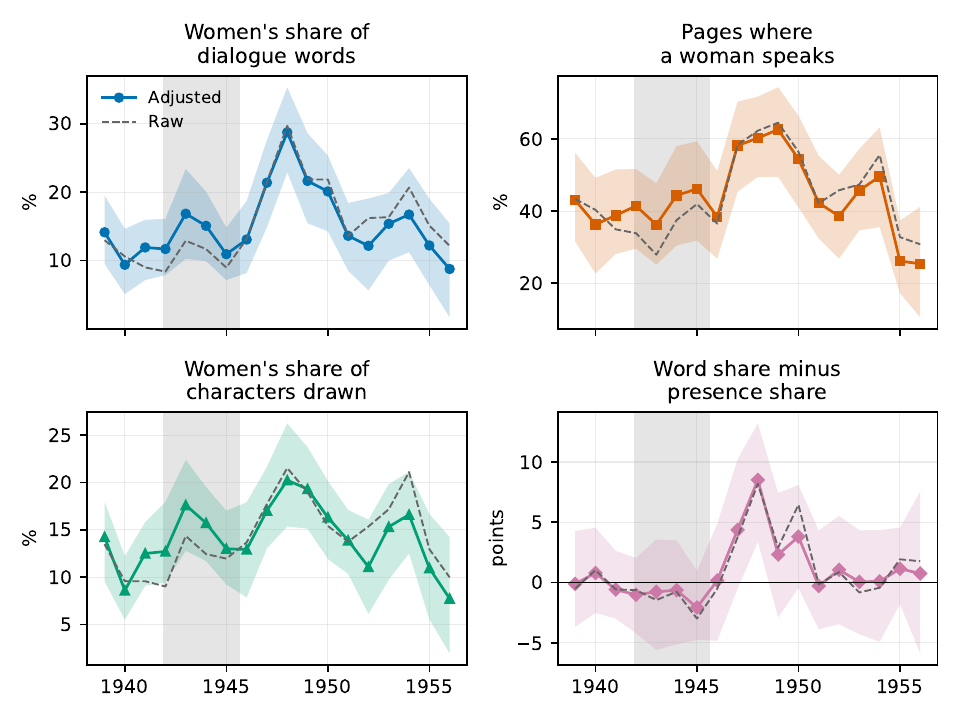}
\caption{Women's voice and presence in American comic books, 1938--1956. Solid lines are
yearly estimates adjusted for publisher group and page genre, with comic-bootstrap 95\% bands;
dashed lines are raw yearly means. The gray band marks the United States at war. Lower right:
women's share of dialogue words minus their share of the characters drawn on the page; zero
means women speak in proportion to their presence.}
\label{fig:voice}
\end{figure}

\paragraph{Findings.} \emph{In American comics, women's share of the dialogue follows their
share of the characters drawn.} Women speak \VWordShare\% of the dialogue words, and a
woman's balloon is as long as a man's (\VWpbF{} against \VWpbM{} words). Across most of the period, women's share of the words on a page moves with their share of the characters drawn, so the low share of dialogue spoken by women arises mainly from the low number of women drawn.
\emph{Women's share of the dialogue rose in the postwar years.} It is \VWordPre\% in 1938--41
and \VWordWar\% in 1942--45, rises to \VWordPost\% in 1946--49 and settles at \VWordFifties\% in
1950--56. Holding publisher and genre fixed, the wartime change (1944--45 against 1940--41) is \VAdjWordWarPre{} points (95\% CI \VAdjWordWarPreLo{} to \VAdjWordWarPreHi), and the change from 1940--41 to 1947--49 is \VAdjWordPostPre{} points (\VAdjWordPostPreLo{} to \VAdjWordPostPreHi); the share falls back by 1953--54 (Figure~\ref{fig:voice}). In the late 1940s women speak more
than their presence alone predicts (adjusted gap \VAdjGapPostPre{} points above 1940--41,
\VAdjGapPostPreLo{} to \VAdjGapPostPreHi). \emph{Women mostly address men, and their share of
the dialogue is highest on pages with a female protagonist.} Of women's balloons, \VToMaleF\%
address a man and \VToFemaleF\% a woman; women speak as the hero's love interest in
\VRoleLoveInterestF\% of their balloons (men \VRoleLoveInterestM\%) and as an authority in
\VRoleAuthorityF\% (men \VRoleAuthorityM\%). On the \VNLead{} pages whose protagonist is coded
as a woman, women speak \VWordLead\% of the words; on the other \VNNoLead{} pages,
\VWordNoLead\%.

\begin{figure}[!htbp]
\centering
\includegraphics[width=\linewidth]{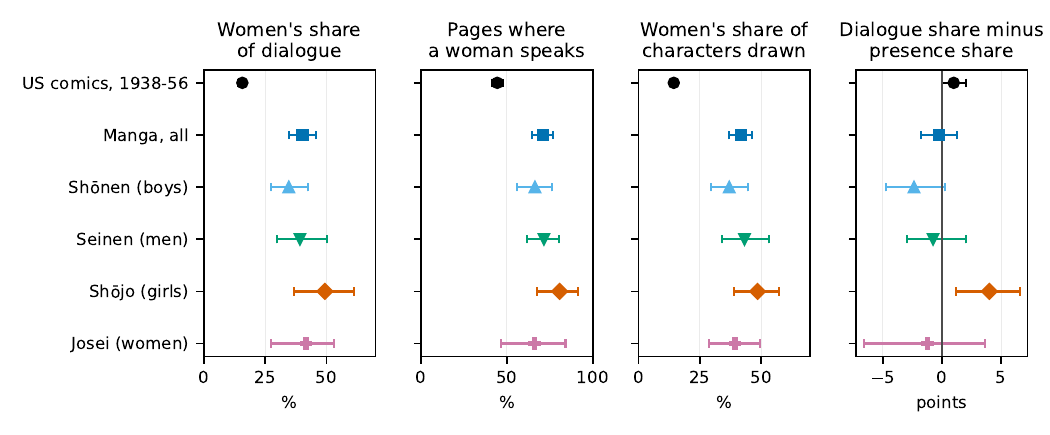}
\caption{Women's voice and presence in American comic books (1938--1956, one random story page
with 3 to 25 dialogue balloons per comic, word shares) and Japanese manga (Manga109-s,
1970s--2010s, \MPages{} interior pages with 3 to 25 text boxes and at least one human speaker
link, shares of text length), overall and by the readership a manga is published for. Points
are estimates with 95\% bootstrap intervals, resampling comics for the American sample and
volumes for manga. Right: women's share of the dialogue minus their share of the characters
drawn. Manga speakers come from the Manga109Dialog human speaker links.}
\label{fig:mangavoice}
\end{figure}

\emph{Women's share of the dialogue is more than twice as high in later manga as in golden-age American comics.} Women speak
\MWordAll\% of the dialogue in manga (95\% CI \MWordAllLo{} to \MWordAllHi), against \MWordUS\%
(\MWordUSLo{} to \MWordUSHi) in American comics of 1938--56 (Figure~\ref{fig:mangavoice}). The difference holds within every readership: \MWordShonen\% in sh\=onen manga written for boys, \MWordSeinen\% in seinen for men, \MWordJosei\% in josei for women and \MWordShojo\% in sh\=ojo for girls. A study of ten series found that male characters speak most of the dialogue in sh\=onen manga and that sh\=ojo is more balanced \citep{unserschutz2015what}, and our estimates extend that finding to four readerships and a whole collection. The two corpora are separated by roughly half a century, American comics of 1938--1956
against manga of the 1970s to the 2010s, so the contrast mixes the difference between the two
traditions with the difference between periods; American comics of the same later decades may
well be more balanced, which a comparison of matched periods would test. As in American comics, women's share of the dialogue
largely tracks their share of the characters on the page (gap \MGapAll{} points in manga,
\MGapUS{} in American comics), so the difference between the two corpora lies mainly in how many women they draw. In sh\=ojo manga, by contrast, women speak more than their presence predicts (\MGapShojo{} points, \MGapShojoLo{} to \MGapShojoHi).

\paragraph{What PAGE enables.} Human speaker links exist for one manga collection and for no American archive, so we use the model to attribute speakers in American comics and the human annotation of manga to check the model's codes, which lets us compare the two traditions with one instrument.

\section{A research agenda for artificial intelligence and comics}\label{sec:discussion}

The withheld face of the sitting president, the wartime change of the page, and the relation between women's dialogue and women's presence each rest on a different capability of PAGE and on thousands of pages, and none could be established without reading every page of an archive. The same instruments open six lines of work for scholarship and for the medium's readers, archives and artists, and for each we state what this paper already provides.

\paragraph{(1) Discovery by what is drawn.} Archives of comics are indexed by title, issue and,
at best, the words on the page. Image embeddings and vision-language reading make them searchable by what is drawn, as Newspaper Navigator did for the photographs and illustrations of historical newspapers \citep{lee2020newspaper}. Study~1 shows the need, because text search reaches pages on which Roosevelt is recognizable only by his face only when another presidential term happens to appear on them.
An image search for unnamed likenesses would let a teacher or a historian assemble every such
page.

\paragraph{(2) Attribution of uncredited artists.} Of the \GStories{} American comic stories of
1938--1956 in the GCD, \GNoPencil\% name no penciller, and most of the rest carry credits
established by later research. Across series, the nearest page is by the same credited
penciller for \StrictUniqueHit\% of pages, \StrictUniqueLift{} times the shuffled baseline.
Embeddings therefore add a visual signal that, combined with signatures, series records and the
judgment of experts, can help credit the artists of the period. The open technical problem is to separate an artist's drawing from recurring characters, which can run in more than one series, and from repeated art.

\paragraph{(3) Access for readers who cannot see the page.} A screen reader or an audio description needs the reading order, speaker links and transcripts that Parse and Annotate already produce \citep{sachdeva2024manga}. Generative models can describe what a panel shows,
and image-aware translation opens a book to readers of another language
\citep{hinami2021towards}. We propose to build and evaluate these tools with blind and low-vision readers.

\paragraph{(4) Preservation and restoration with provenance.} The checksummed, indexed mirror we built for this study already keeps the COMICS scans retrievable. Generative restoration and colorization
\citep{furusawa2017comicolorization} can produce readable access copies of damaged pages,
provided each copy is kept apart from the original scan and labeled with how it was made.
Archives will also need to detect synthetic pages that enter their collections \citep{ferrara2026paradox}, a task in which expert readers and automated detectors work best together \citep{ha2024organic}.

\paragraph{(5) A comparative science of the medium.} Theories of comics as a visual language
with national dialects \citep{cohn2013visual,cohn2011different} become testable across whole
traditions once pages from several of them are parsed and annotated with one instrument. Study~3
compares American comics and manga; extending the comparison to matched periods, and to
European comics and webcomics, would turn the present contrasts into tests of how traditions
diverge and influence one another.

\paragraph{(6) Shared instruments and human references.} Model readings become cumulative
evidence when their codebooks, annotations and human references are shared. The gender-named characters of Manga109 served here as a human reference that required no new annotation; deliberate
reference sets for narrative role, depiction and speaker identity, and replication of each
codebook across model families, would let every future study report its error against the same standard. The fan communities that assembled these archives are natural partners in building such references \citep{jenkins2006convergence}.

\paragraph{Risks and how the framework measures them.} A model that reads comics can make
systematic errors that two of its own sessions will reproduce. It can carry the demographic and cultural biases of its training \citep{ferrara2023chatgpt} and change its answers with the wording of a prompt \citep{ye2026robustness}. We measure these errors on samples checked against independent evidence (Section~\ref{sec:grounding}), including the human reference for speaker gender in manga. Generative AI also poses risks to the people who make comics, through imitation of their styles and training on their work without consent \citep{jiang2023ai,epstein2023art}, and to archives, through synthetic pages that blur provenance \citep{ferrara2026paradox}; artists can protect their work against the first with style cloaking and data poisoning \citep{shan2023glaze,shan2024nightshade}. Every use we propose is analytic or serves access, none generates new pages in an artist's
style, and attribution remains the decision of human experts.

\paragraph{Limitations.} Gender is coded as each character is drawn and presented in the story,
not as any real identity. The presidents study has no second coder; the same model family reviewed its likeness-based identifications in a second reading of the page image, these identifications await a human check, and its discovery reaches only pages whose OCR names a president. All codebook annotation uses one model family, and the human references cover speaker gender in manga, mostly for minor characters; narrative role, addressee and tone have no human reference, nor do the American pages, whose balloons come from noisy OCR and whose speaker sample draws one random page per comic.
Duplicate scans were searched only among the comics of Study~1, and selection within publishers
and genres cannot be ruled out by fixed effects. The manga comparison spans different decades
(the 1970s to the 2010s against 1938--1956), and Manga109-s is a curated set of \MVols{}
volumes, so we draw no causal conclusion from the contrast between the two traditions. Penciller matching cannot exclude recurring features that run in more than one series. COMICS is a selection of public-domain books, and its reconstructed dates and GCD links carry noise
that biases tests such as penciller matching toward the null.

\section{Conclusions}\label{sec:conclusions}

Comics are a visual record that a mass readership lived with for generations, and most of that
record has never been studied as a whole. We proposed PAGE, a framework that parses pages into structured objects, annotates them with a vision-language model under written codebooks, grounds samples of the model's codes in human annotation and independent checks, and explains change across an archive with publisher and genre held fixed. Three studies of \NGrammarComics{} American comic books and \MVols{} volumes of manga, each built on a different capability, find that golden-age comics withheld the sitting president's face, that the American page carried fewer and larger panels in wartime and more words after it, and that in both traditions women's share of the dialogue largely tracks their share of the characters drawn. The same instruments can make archives searchable
by what is drawn, open comics to blind and low-vision readers, help preserve fragile
collections and help credit the artists of the period. Grounded in human annotation, and used with the consent and credit of the people who make comics, artificial intelligence can extend the study of the medium to whole archives and serve its readers, archives and artists.

\backmatter

\bmhead{Acknowledgements}
We thank the creators of the COMICS corpus, the Digital Comic Museum, the Grand Comics Database
volunteers, and the Manga109 and Manga109Dialog teams at the University of Tokyo. Manga109-s is used under its research license, which permits reproducing up to 20\% of the pages of each volume with credit to the author; the four manga pages in Appendix Figure~\ref{fig:mangapages} follow these terms.

\section*{Declarations}

\begin{itemize}
\item \textbf{Availability of data and materials.} COMICS is available from its authors
\citep{iyyer2017amazing}; Manga109-s and Manga109Dialog are available from the University of
Tokyo under their licenses \citep{aizawa2020building,li2024manga109dialog}; the GCD is available at \url{https://www.comics.org} \citep{gcd2026}. The code, codebooks, annotations and derived measures will be
released in a public repository upon publication; page images are not redistributed.
\item \textbf{Use of large language models.} Claude Opus 5.5 (Anthropic) was the annotation
instrument described in Section~\ref{sec:framework}: it coded the presidents pages, the speaker
pages and the manga characters, read publication years and titles from covers and indicia, and
judged retrieval relevance in Section~\ref{sec:grounding} (OpenAI's \texttt{gpt-6-astra}, run through the Codex CLI,
judged a subset of \XJudgePages{} pages, with $\kappa=\XJudgeKappa$ between the two). Claude
also assisted with code, reference retrieval and drafting under the author's direction; the
author reviewed all outputs and takes responsibility for the content.
\item \textbf{Ethics approval and consent to participate.} Not applicable.
\item \textbf{Competing interests.} The author declares no competing interests.
\item \textbf{Funding.} Not applicable.
\item \textbf{Authors' contributions.} EF conceived the study, designed the framework, directed
the analyses and wrote the manuscript.
\end{itemize}

\begin{appendices}
\section{Data notes}\label{app:data}

\paragraph{Page numbering in the panel-annotation subset of COMICS.} The widely used
panel-annotation subset numbers its page images differently from the full archive. When we
match all \XPackage{} images by checksum to the byte-identical page in the full scans,
\XOffset{} carry a different page key, almost always one page higher. The OCR follows the full
archive, so a study that joins the subset's images to the OCR by key reads the wrong page's text
for about half the pages; all results in this article use the full archive.

\begin{figure}[p]\centering
\begin{subfigure}[t]{0.24\linewidth}\centering\includegraphics[width=1\linewidth,height=0.3\textheight,keepaspectratio]{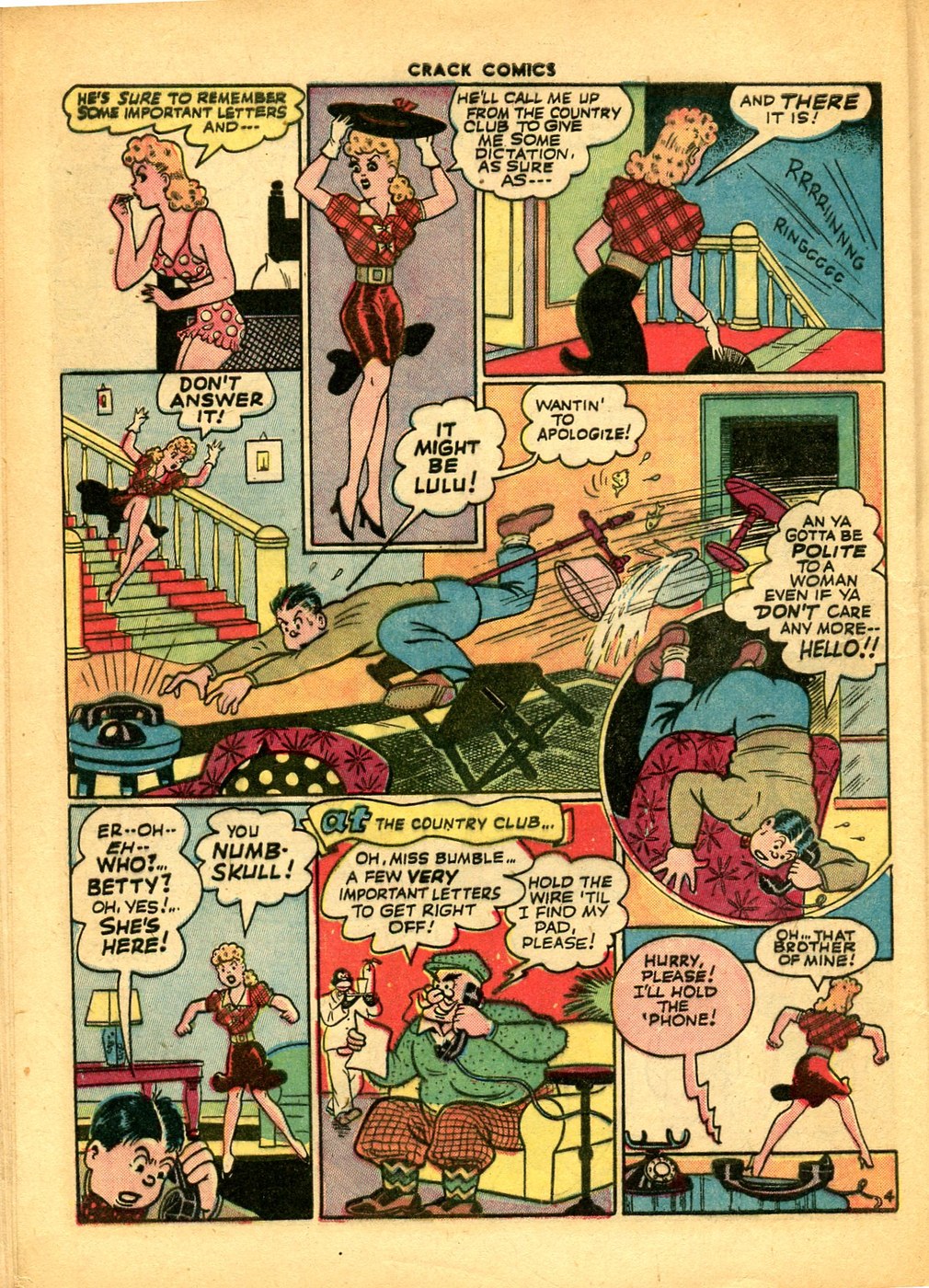}\caption{\emph{Crack Comics} \#32, Quality, 1943, \texttt{3496\_35}.}\end{subfigure}
\hfill
\begin{subfigure}[t]{0.24\linewidth}\centering\includegraphics[width=1\linewidth,height=0.3\textheight,keepaspectratio]{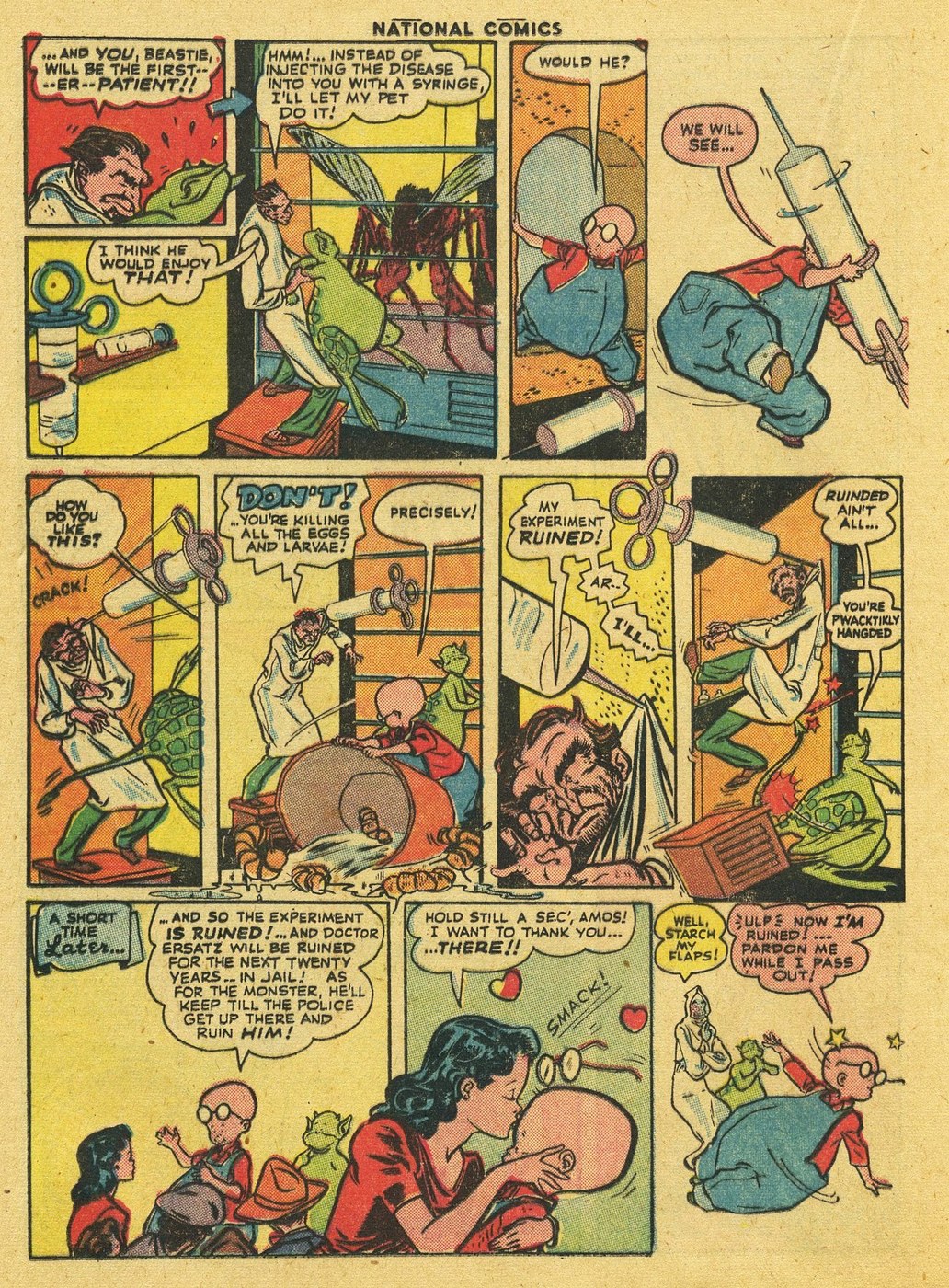}\caption{\emph{National Comics} \#50, Quality, 1945, \texttt{2424\_35}.}\end{subfigure}
\hfill
\begin{subfigure}[t]{0.24\linewidth}\centering\includegraphics[width=1\linewidth,height=0.3\textheight,keepaspectratio]{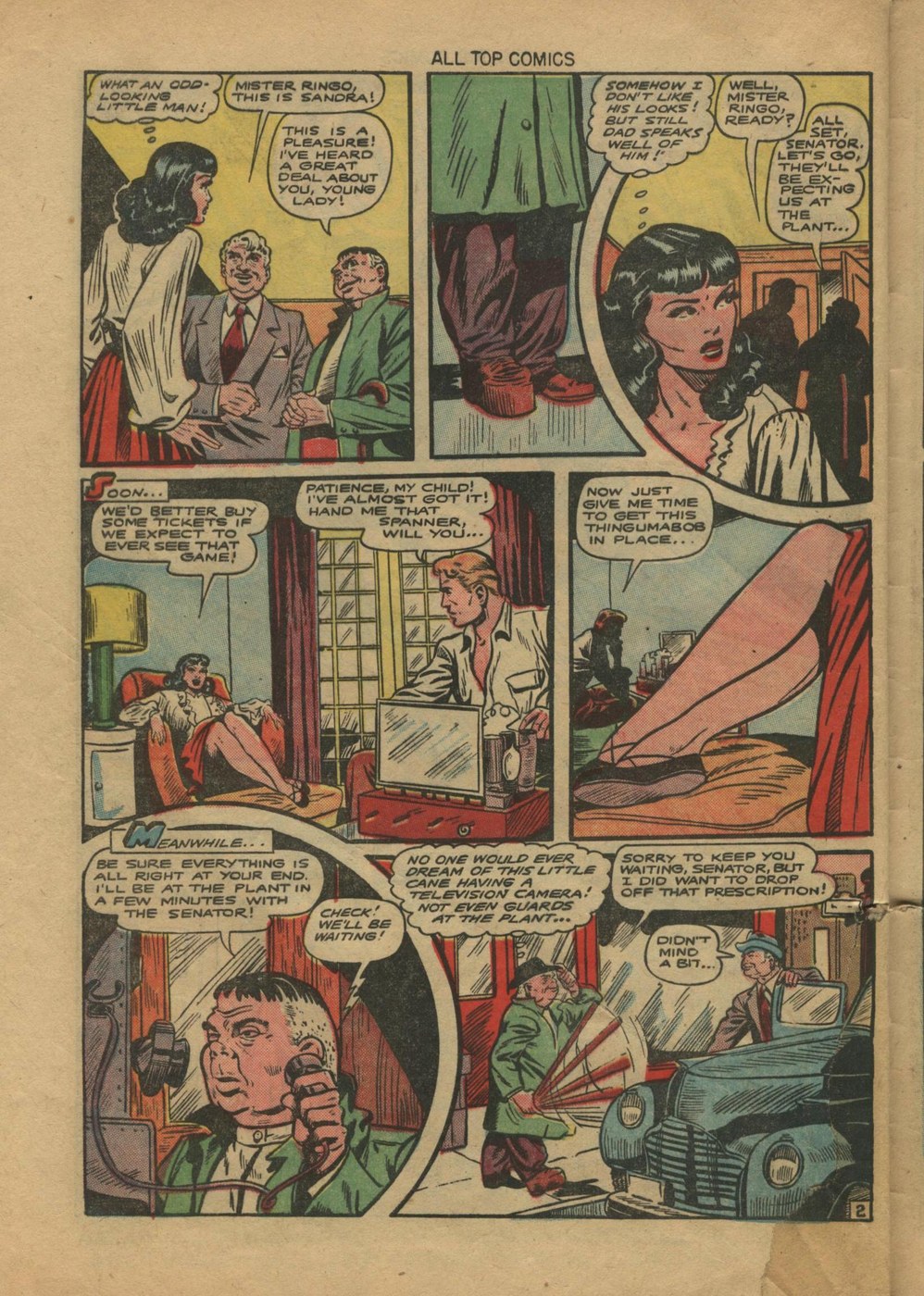}\caption{\emph{All Top Comics} \#10, Fox, 1948, \texttt{163\_27}.}\end{subfigure}
\hfill
\begin{subfigure}[t]{0.24\linewidth}\centering\includegraphics[width=1\linewidth,height=0.3\textheight,keepaspectratio]{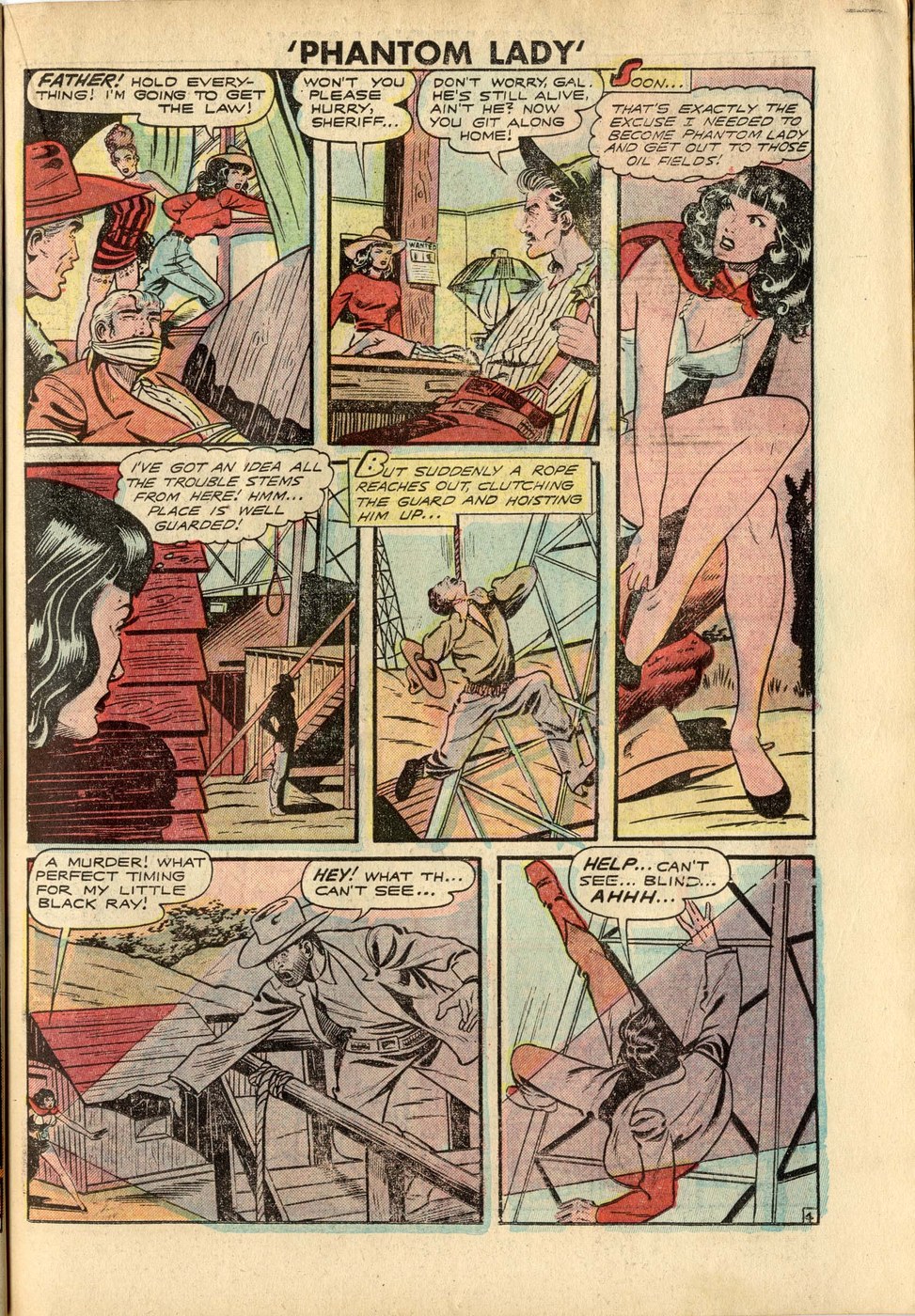}\caption{\emph{Phantom Lady} \#20, Fox, 1948, \texttt{2173\_14}.}\end{subfigure}
\hfill
\par\medskip
\begin{subfigure}[t]{0.24\linewidth}\centering\includegraphics[width=1\linewidth,height=0.3\textheight,keepaspectratio]{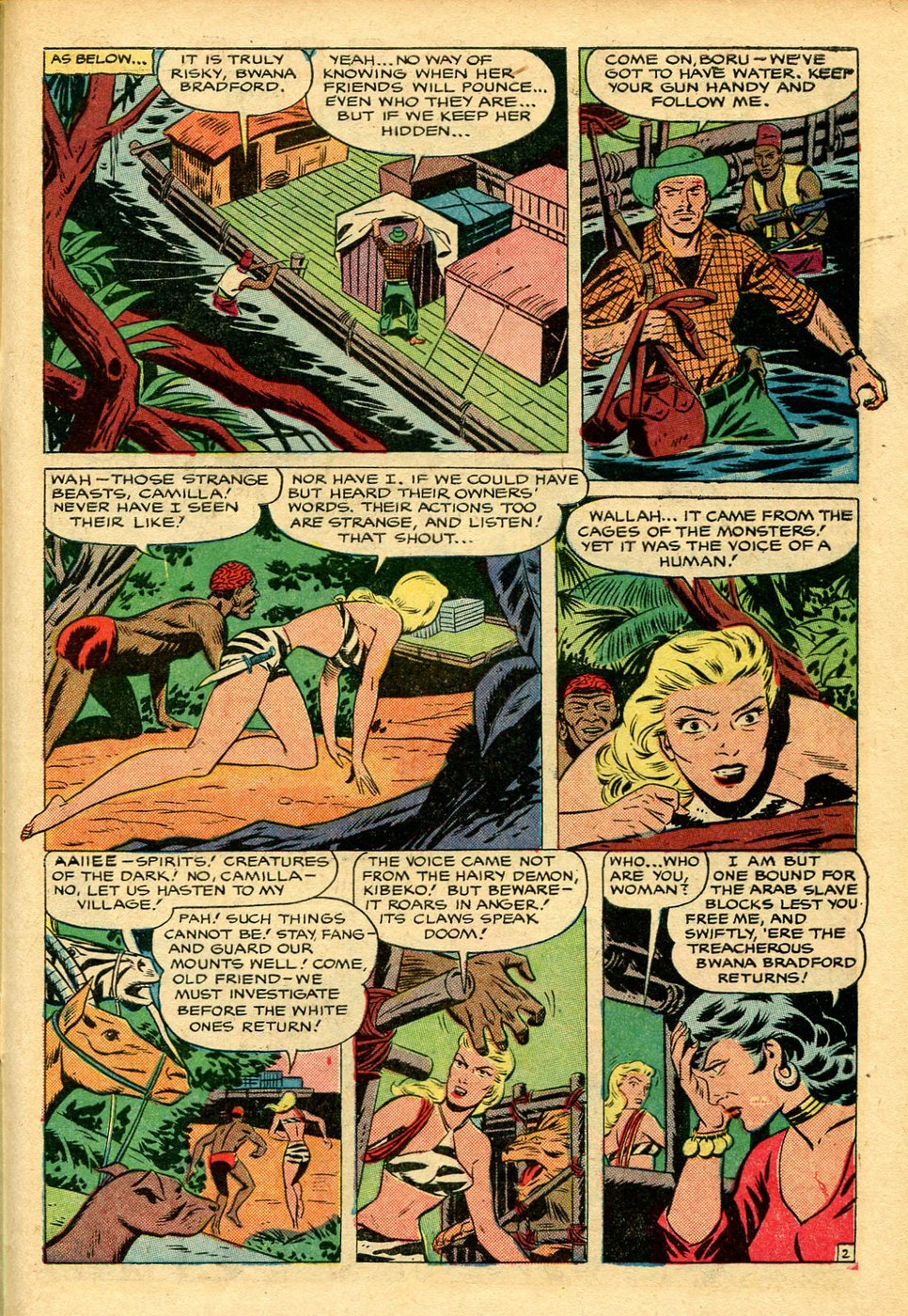}\caption{\emph{Jungle Comics} \#131, Fiction House, 1950, \texttt{3683\_40}.}\end{subfigure}
\hfill
\begin{subfigure}[t]{0.24\linewidth}\centering\includegraphics[width=1\linewidth,height=0.3\textheight,keepaspectratio]{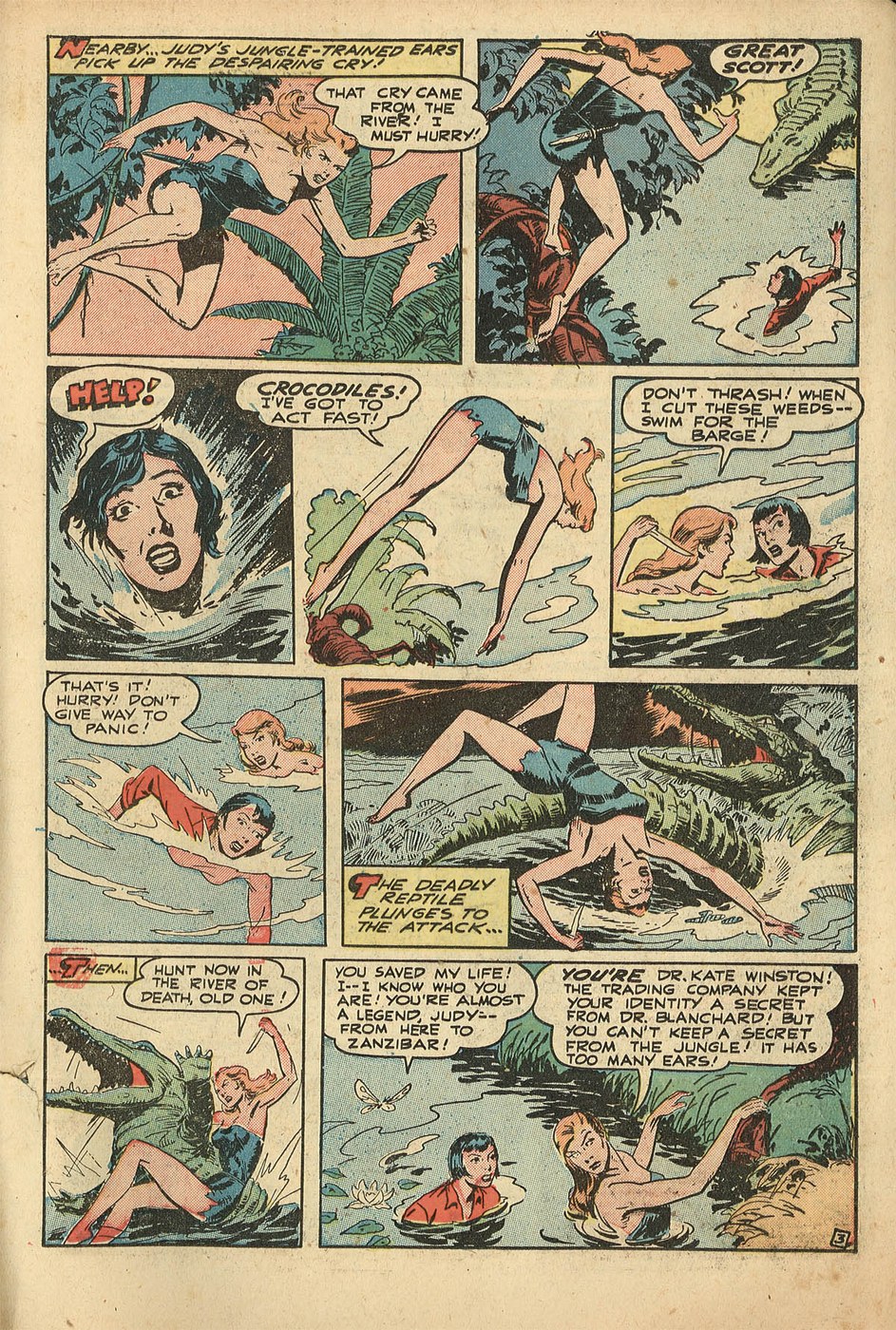}\caption{\emph{Exciting Comics} \#68, Nedor/Standard, 1949, \texttt{1088\_14}.}\end{subfigure}
\hfill
\begin{subfigure}[t]{0.24\linewidth}\centering\includegraphics[width=1\linewidth,height=0.3\textheight,keepaspectratio]{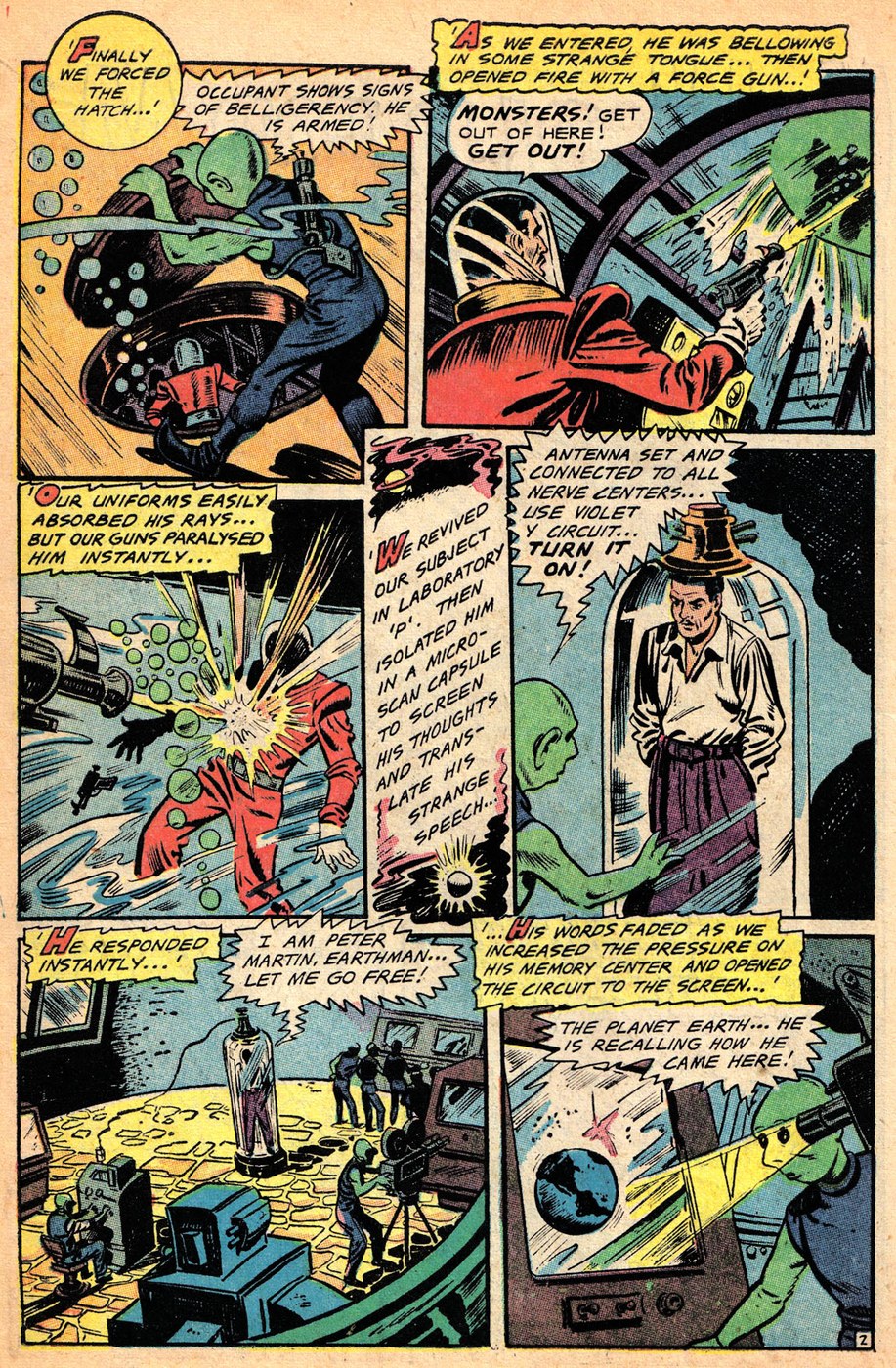}\caption{\emph{Ghost Comics} \#4, Fiction House, 1952, \texttt{8\_9}.}\end{subfigure}
\hfill
\begin{subfigure}[t]{0.24\linewidth}\centering\includegraphics[width=1\linewidth,height=0.3\textheight,keepaspectratio]{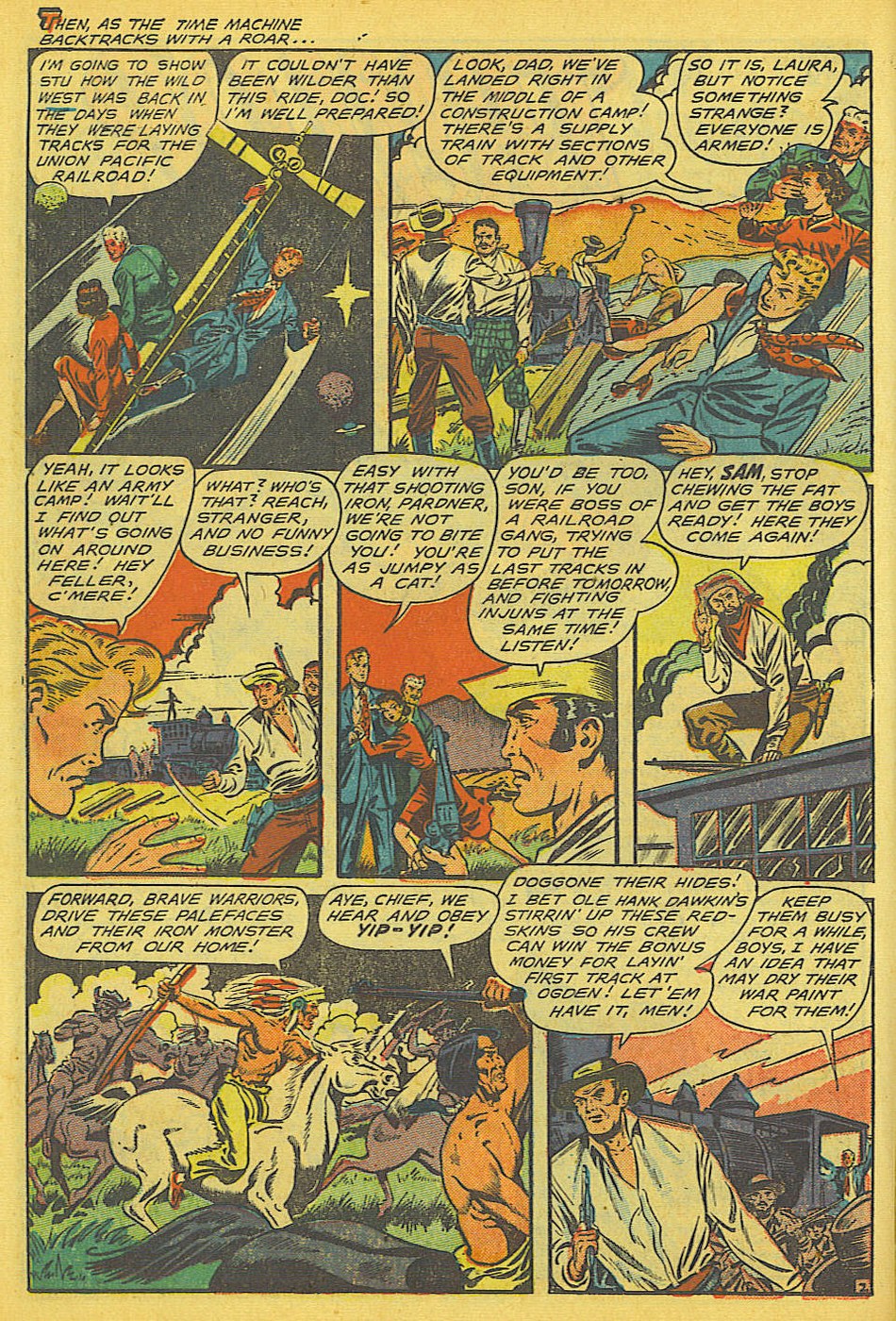}\caption{\emph{Jumbo Comics} \#131, Fiction House, 1949, \texttt{1580\_21}.}\end{subfigure}
\hfill
\par\medskip
\caption{Study~2: pages matched to their nearest neighbor in another series by the same credited penciller: Bernard Dibble (a--b), Matt Baker (c--d), Ralph Mayo (e--f), Jack Kamen (g--h). Pairs are the most similar such matches after excluding reprints and repeated pages (image cosine $\geq 0.95$ to a page of another comic, or shared lettering).}\label{fig:pairs}\end{figure}

\paragraph{Duplicate scans.} Some printed issues appear twice in COMICS as separate scans (for
example \emph{Liberty Scouts Comics} \#2 as comics 1744 and 1745). Among the
\NCandComicsCorpus{} comics read for Study~1 we find \NDupComics{} such pairs by perceptual hash,
confirm each by inspection, and count each issue once in every study. Pages that reprint the
same art in another series, and advertisements repeated across comics, are excluded from the
penciller-matching test in Study~2 (Appendix~\ref{app:gallery}, Figure~\ref{fig:pairs}).

\section{Gallery of pages}\label{app:gallery}

COMICS pages in this gallery are public-domain scans from the Digital
Comic Museum, reproduced unaltered and identified by their page key in the COMICS release.
Manga pages are from Manga109-s, reproduced under its license (one page per volume, courtesy
of the author) \citep{matsui2016sketch,aizawa2020building}, with speakers from Manga109Dialog
\citep{li2024manga109dialog}.

\begin{figure}[p]\centering
\begin{subfigure}[t]{0.48\linewidth}\centering\includegraphics[width=\linewidth,height=0.32\textheight,keepaspectratio]{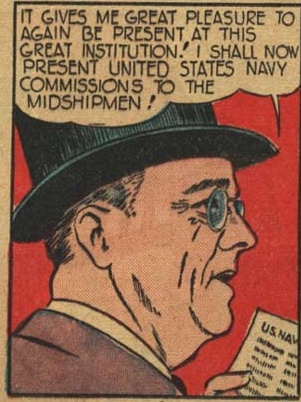}\caption{\emph{Pep Comics} \#16 (M.L.J., June 1941), \texttt{2124\_45}. The president presenting Navy commissions is never named; the pince-nez and the features point to Franklin Roosevelt.}\end{subfigure}
\hfill
\begin{subfigure}[t]{0.48\linewidth}\centering\includegraphics[width=\linewidth,height=0.32\textheight,keepaspectratio]{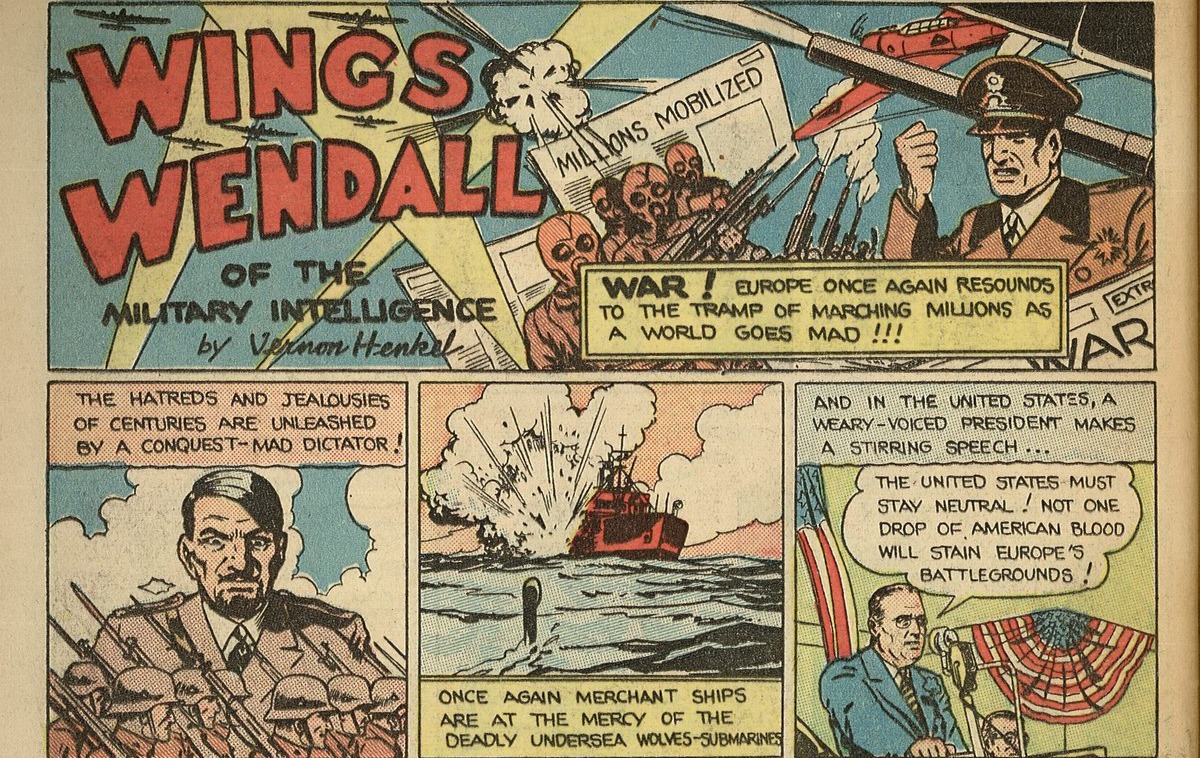}\caption{\emph{Smash Comics} \#6 (Quality, January 1940), \texttt{2630\_62}. An unnamed, ``weary-voiced'' president pledges neutrality at the microphone.}\end{subfigure}
\par\medskip
\begin{subfigure}[t]{0.48\linewidth}\centering\includegraphics[width=\linewidth,height=0.32\textheight,keepaspectratio]{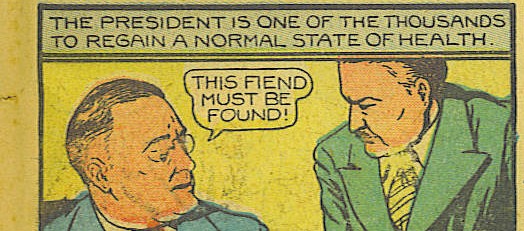}\caption{\emph{Big 3} \#2 (Fox, Winter 1941), \texttt{383\_59}. The recovered president demands that ``this fiend must be found''.}\end{subfigure}
\hfill
\begin{subfigure}[t]{0.48\linewidth}\centering\includegraphics[width=\linewidth,height=0.32\textheight,keepaspectratio]{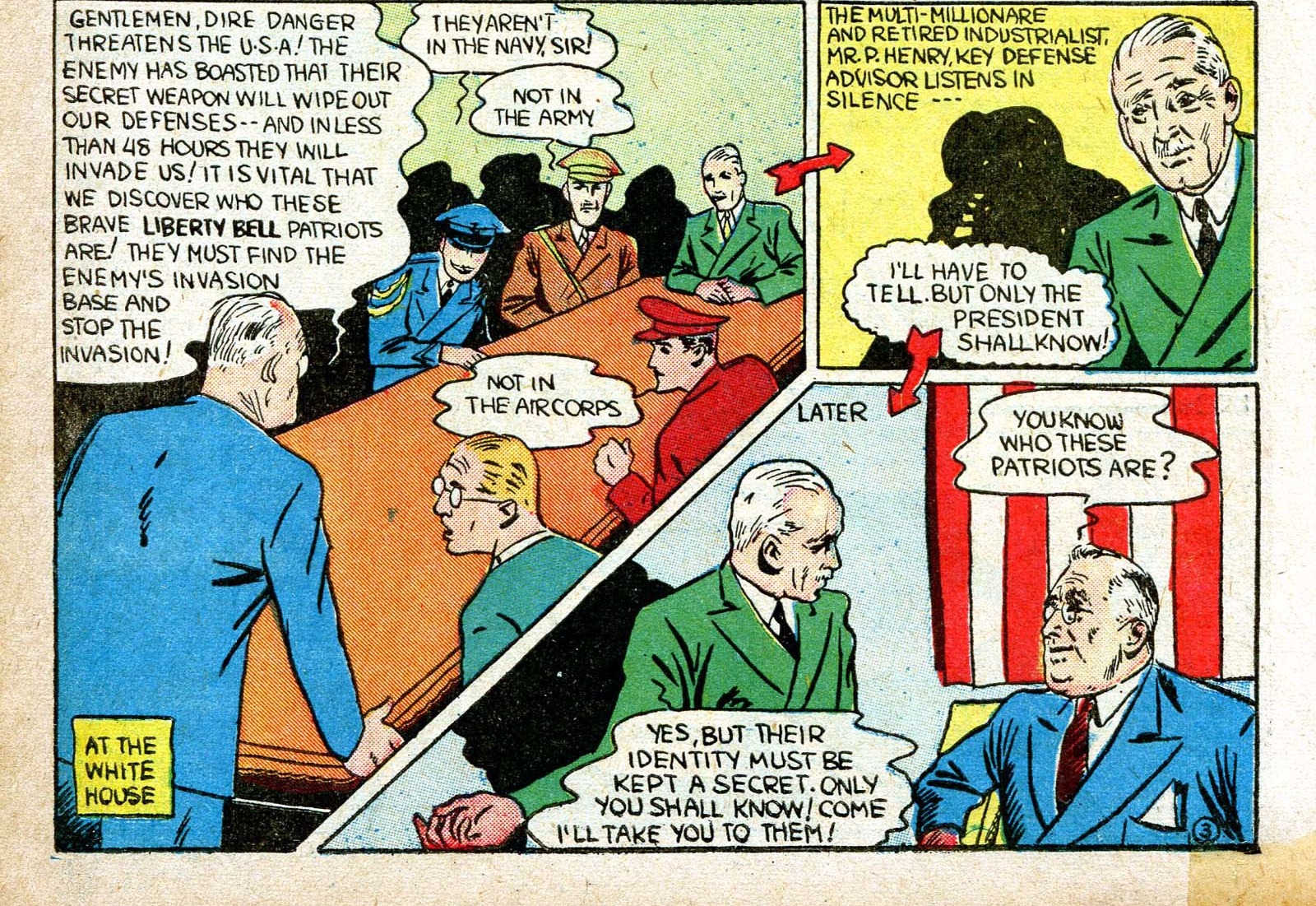}\caption{\emph{Liberty Scouts Comics} \#2 (Comic Corporation of America, June 1941), \texttt{1745\_4}. The president, seen from behind at the White House and then at his desk, is never named.}\end{subfigure}
\par\medskip
\caption{Study~1: the President in golden-age comics. Unaltered crops of the original scans; titles and dates are read from each comic's cover and indicia.}\label{fig:plates}\end{figure}
\begin{figure}[p]\centering
\begin{subfigure}[t]{0.48\linewidth}\centering\includegraphics[width=\linewidth,height=0.32\textheight,keepaspectratio]{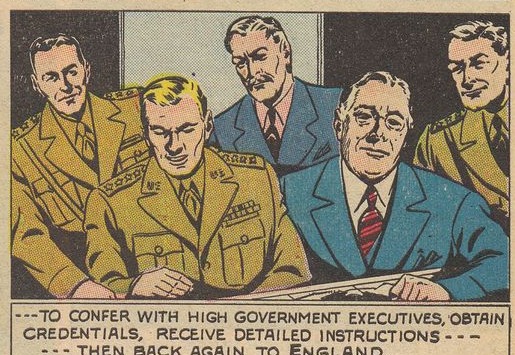}\caption{\emph{War Heroes} \#4 (Dell, April--June 1943), \texttt{3150\_4}. The president, unnamed, with Generals Clark and Eisenhower; the caption says only ``high government executives''.}\end{subfigure}
\hfill
\begin{subfigure}[t]{0.48\linewidth}\centering\includegraphics[width=\linewidth,height=0.32\textheight,keepaspectratio]{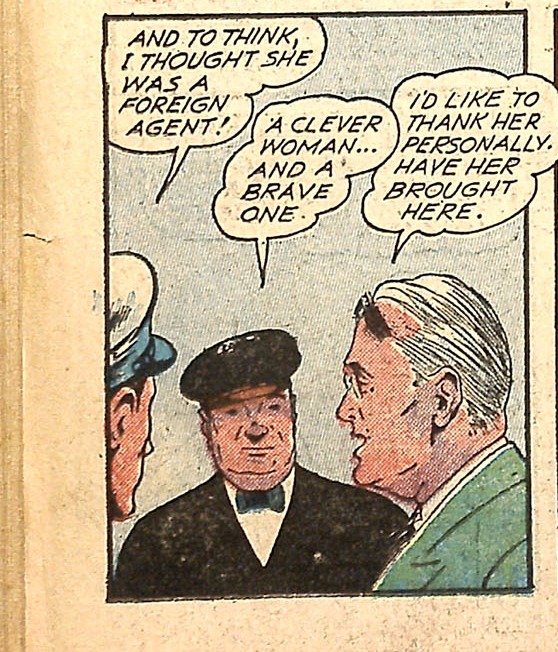}\caption{\emph{Dynamic Comics} \#3 (Chesler, year not printed), \texttt{1039\_34}. Churchill and the president after a foiled assassination.}\end{subfigure}
\par\medskip
\begin{subfigure}[t]{0.48\linewidth}\centering\includegraphics[width=\linewidth,height=0.32\textheight,keepaspectratio]{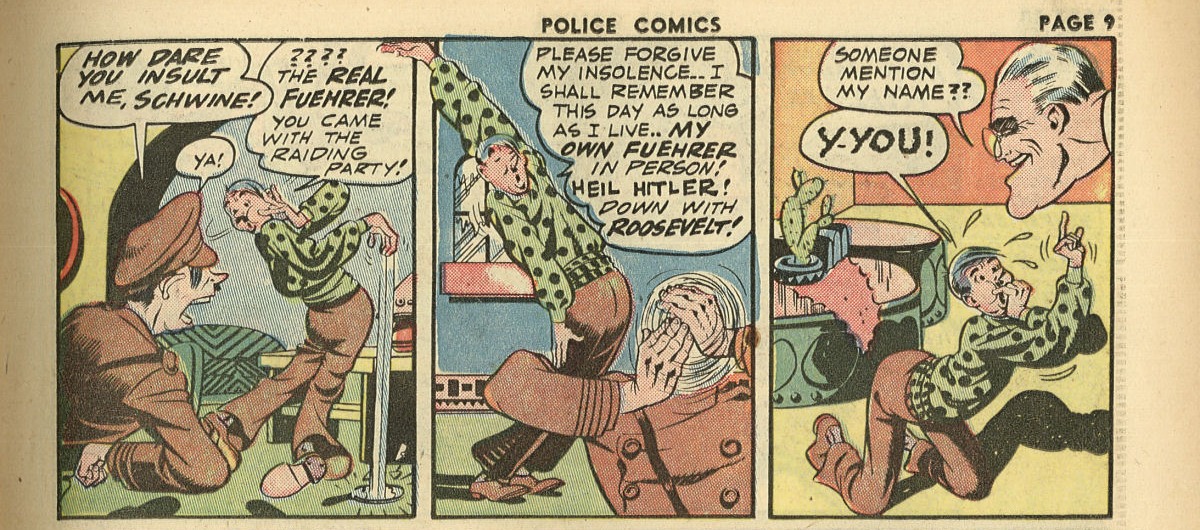}\caption{\emph{Police Comics} \#15 (Quality, January 1943), \texttt{2265\_10}. A Nazi shouts ``Down with Roosevelt!''; the face is most likely Plastic Man's disguise.}\end{subfigure}
\hfill
\begin{subfigure}[t]{0.48\linewidth}\centering\includegraphics[width=\linewidth,height=0.32\textheight,keepaspectratio]{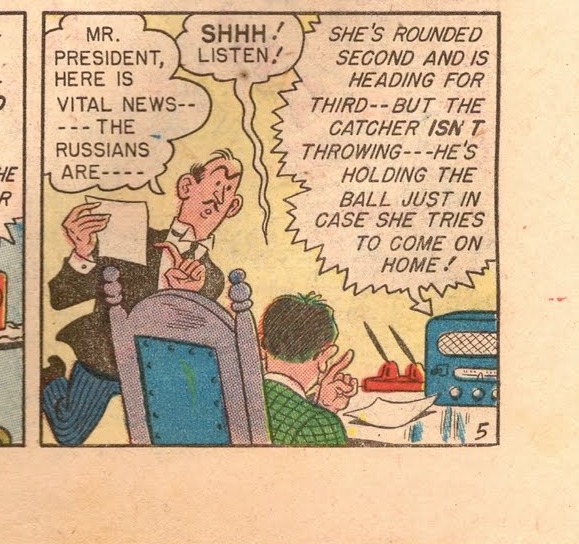}\caption{\emph{Babe} \#3 (Prize, October--November 1948), \texttt{293\_36}. One of the three references that mock a real president: he hushes an aide to follow a baseball game.}\end{subfigure}
\par\medskip
\caption{Study~1: the President in golden-age comics (continued). Unaltered crops of the original scans; titles and dates are read from each comic's cover and indicia.}\label{fig:plates2}\end{figure}
\begin{figure}[p]\centering
\begin{subfigure}[t]{0.32\linewidth}\centering\includegraphics[width=1\linewidth,height=0.3\textheight,keepaspectratio]{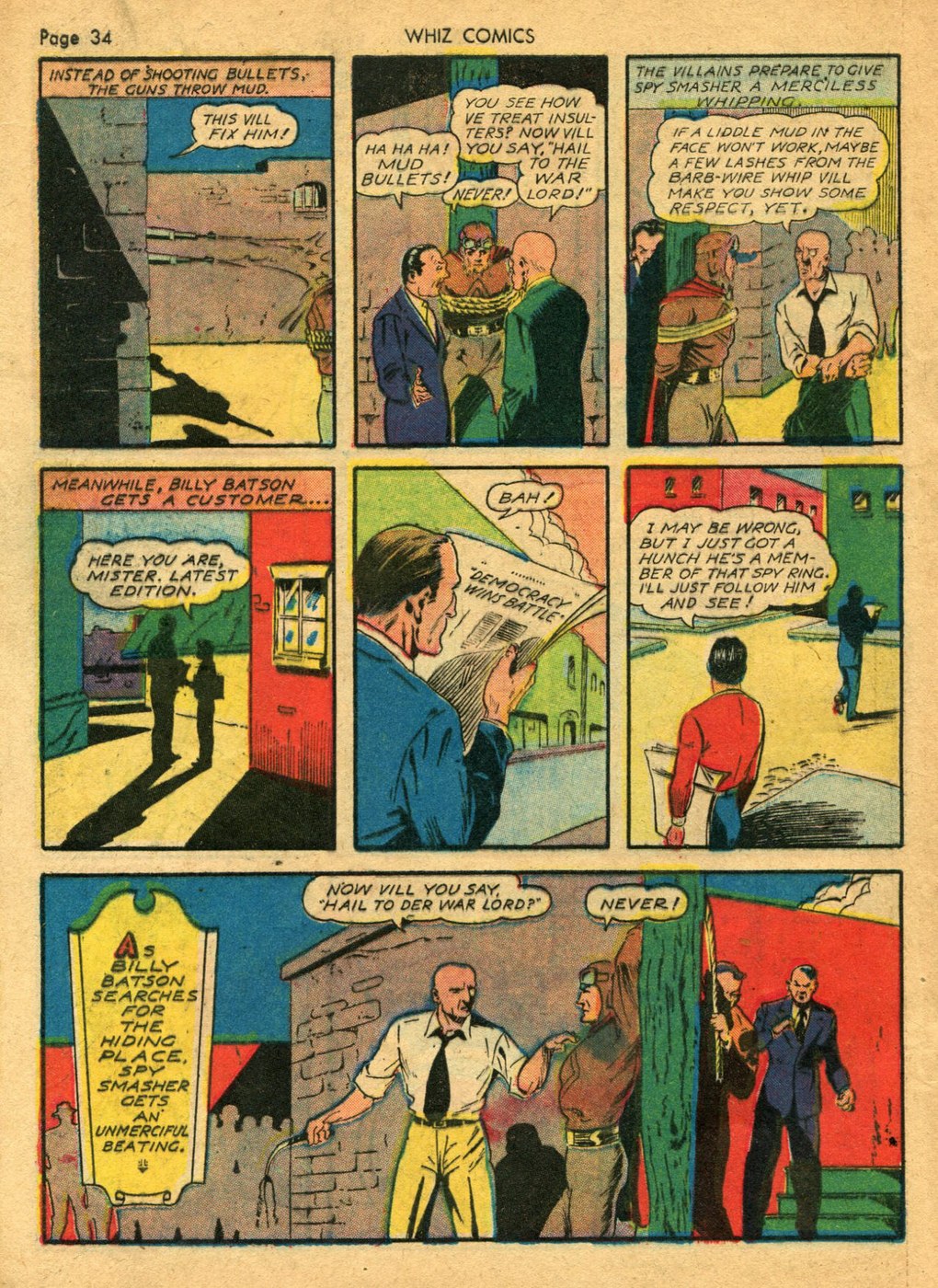}\caption{1940--41: \emph{Whiz Comics} \#18, Fawcett, 1941, \texttt{3934\_35}; 8 panels, 142 words.}\end{subfigure}
\hfill
\begin{subfigure}[t]{0.32\linewidth}\centering\includegraphics[width=1\linewidth,height=0.3\textheight,keepaspectratio]{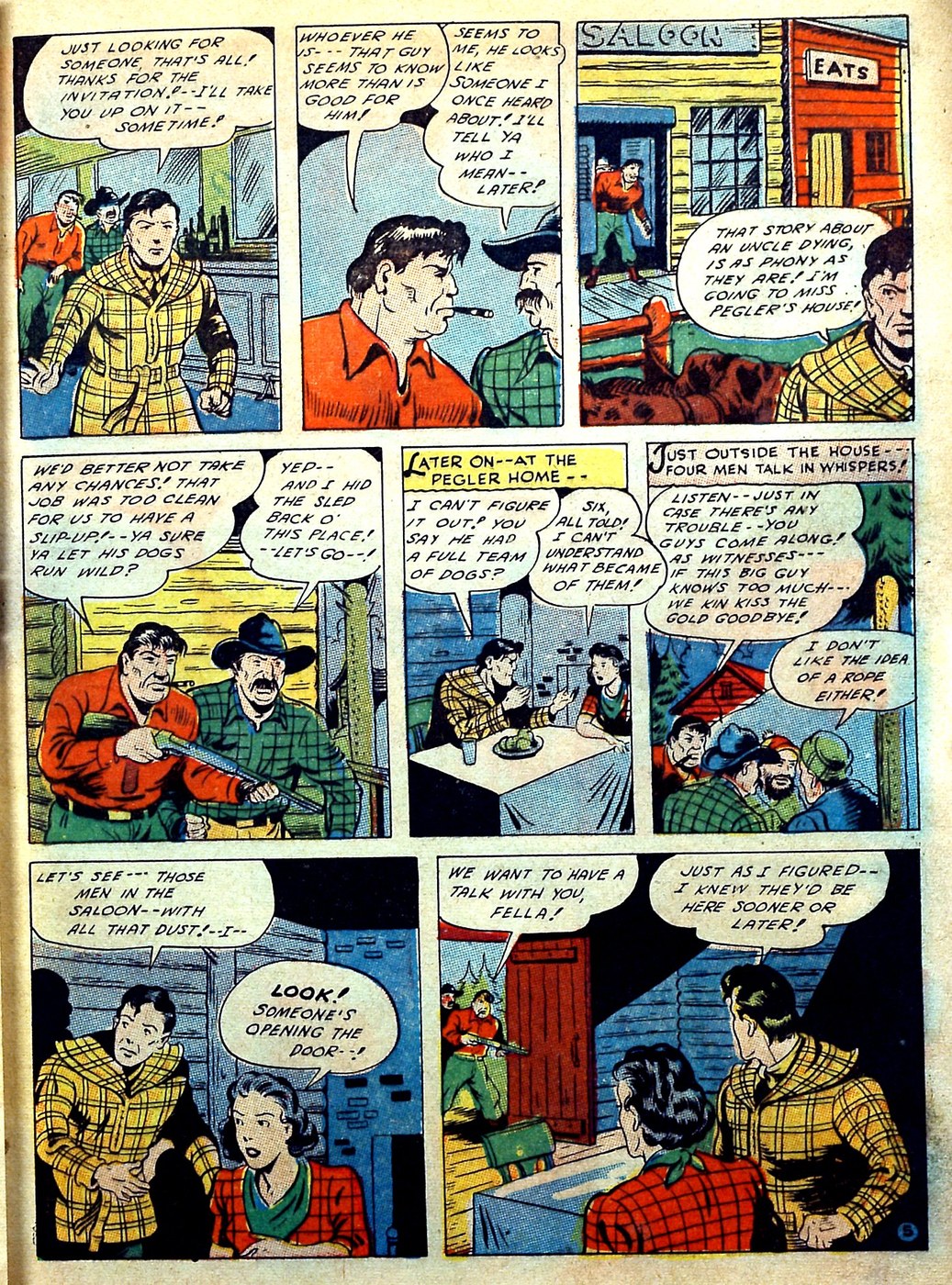}\caption{1940--41: Title not matched, 1941, \texttt{2903\_15}; 8 panels, 148 words.}\end{subfigure}
\hfill
\begin{subfigure}[t]{0.32\linewidth}\centering\includegraphics[width=1\linewidth,height=0.3\textheight,keepaspectratio]{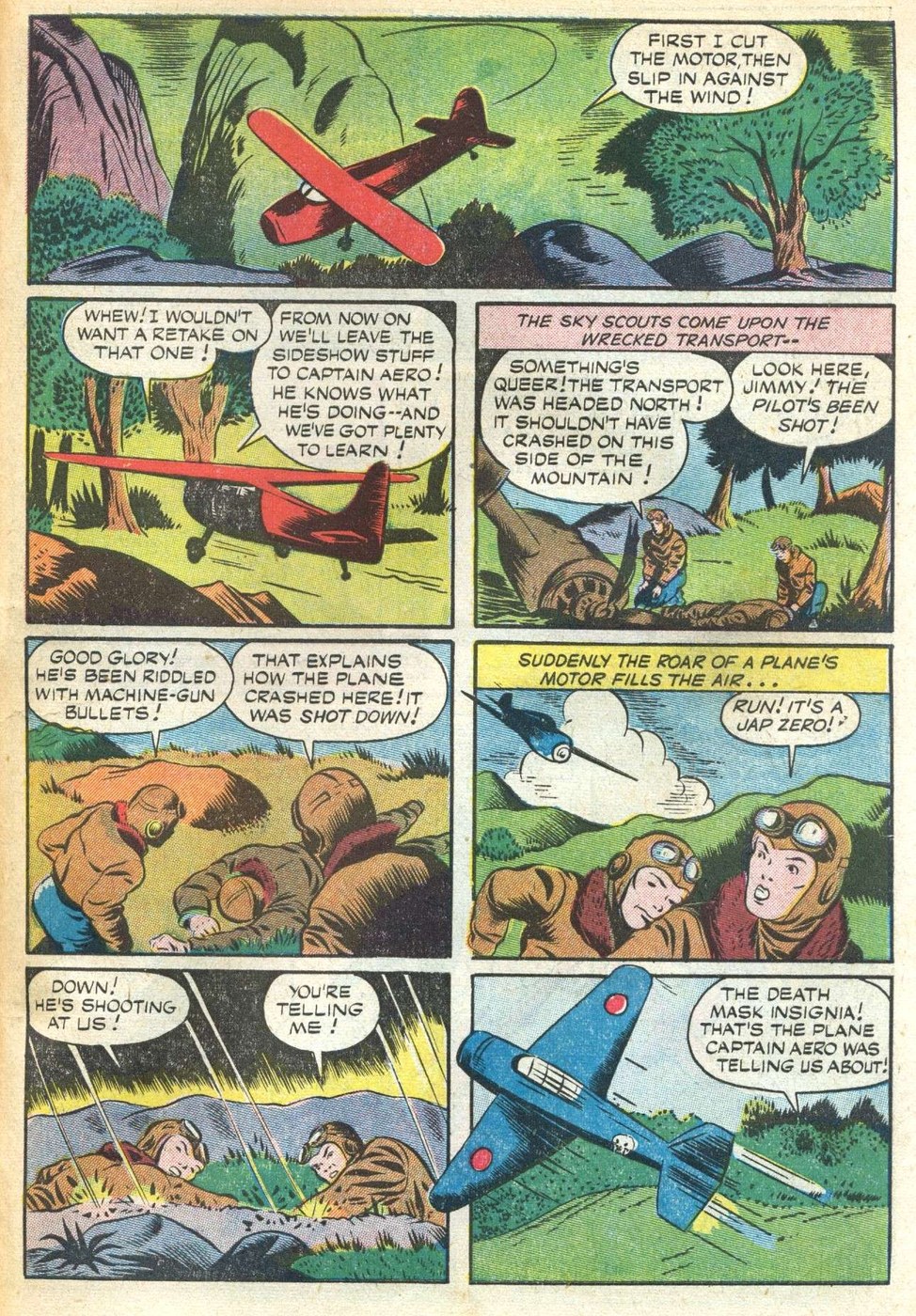}\caption{1944--45: \emph{Captain Aero Comics} \#3 [17], 1944, \texttt{624\_46}; 7 panels, 158 words.}\end{subfigure}
\par\medskip
\begin{subfigure}[t]{0.32\linewidth}\centering\includegraphics[width=1\linewidth,height=0.3\textheight,keepaspectratio]{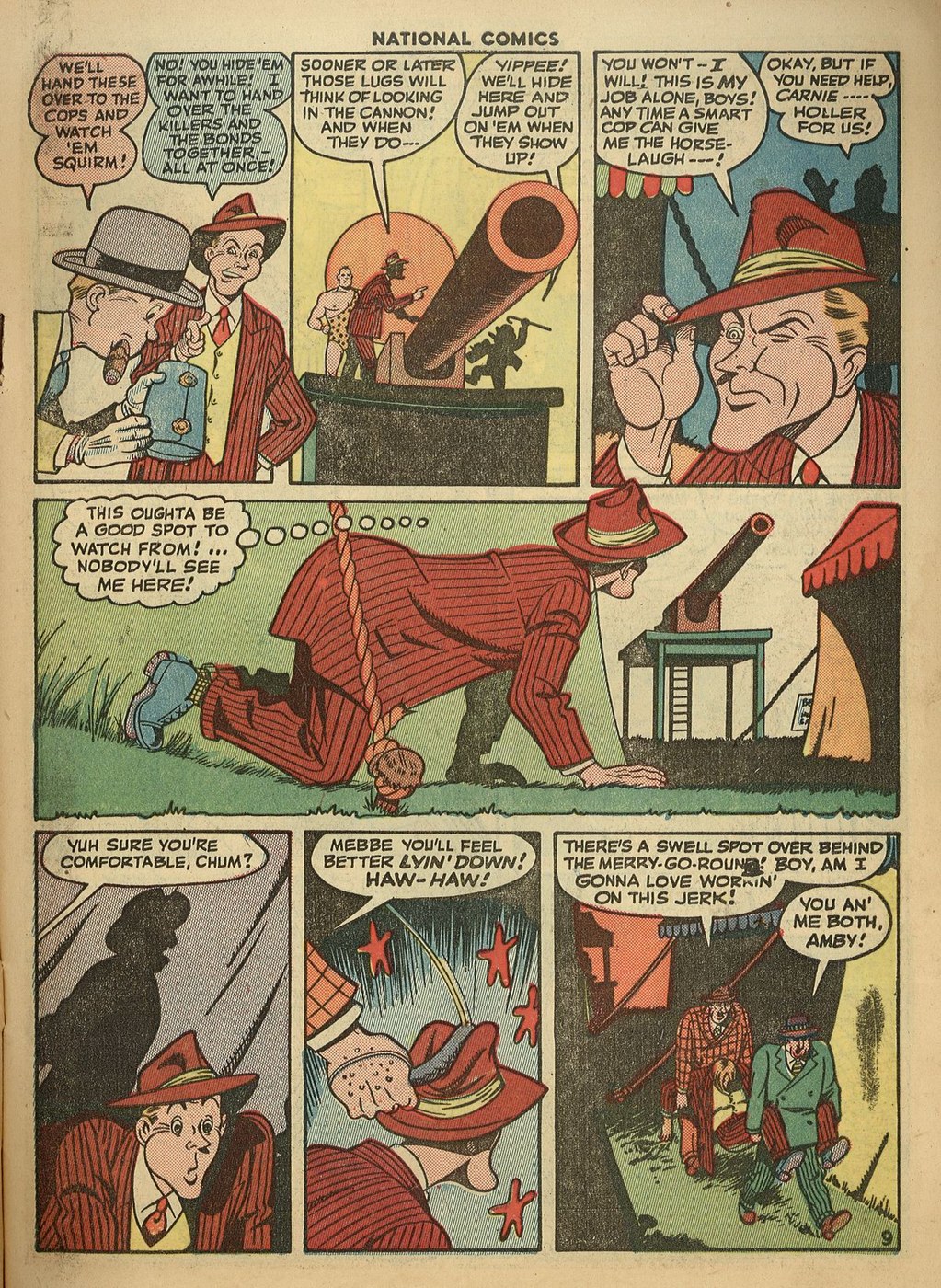}\caption{1944--45: \emph{National Comics} \#49, Quality, 1945, \texttt{1996\_10}; 7 panels, 161 words.}\end{subfigure}
\hfill
\begin{subfigure}[t]{0.32\linewidth}\centering\includegraphics[width=1\linewidth,height=0.3\textheight,keepaspectratio]{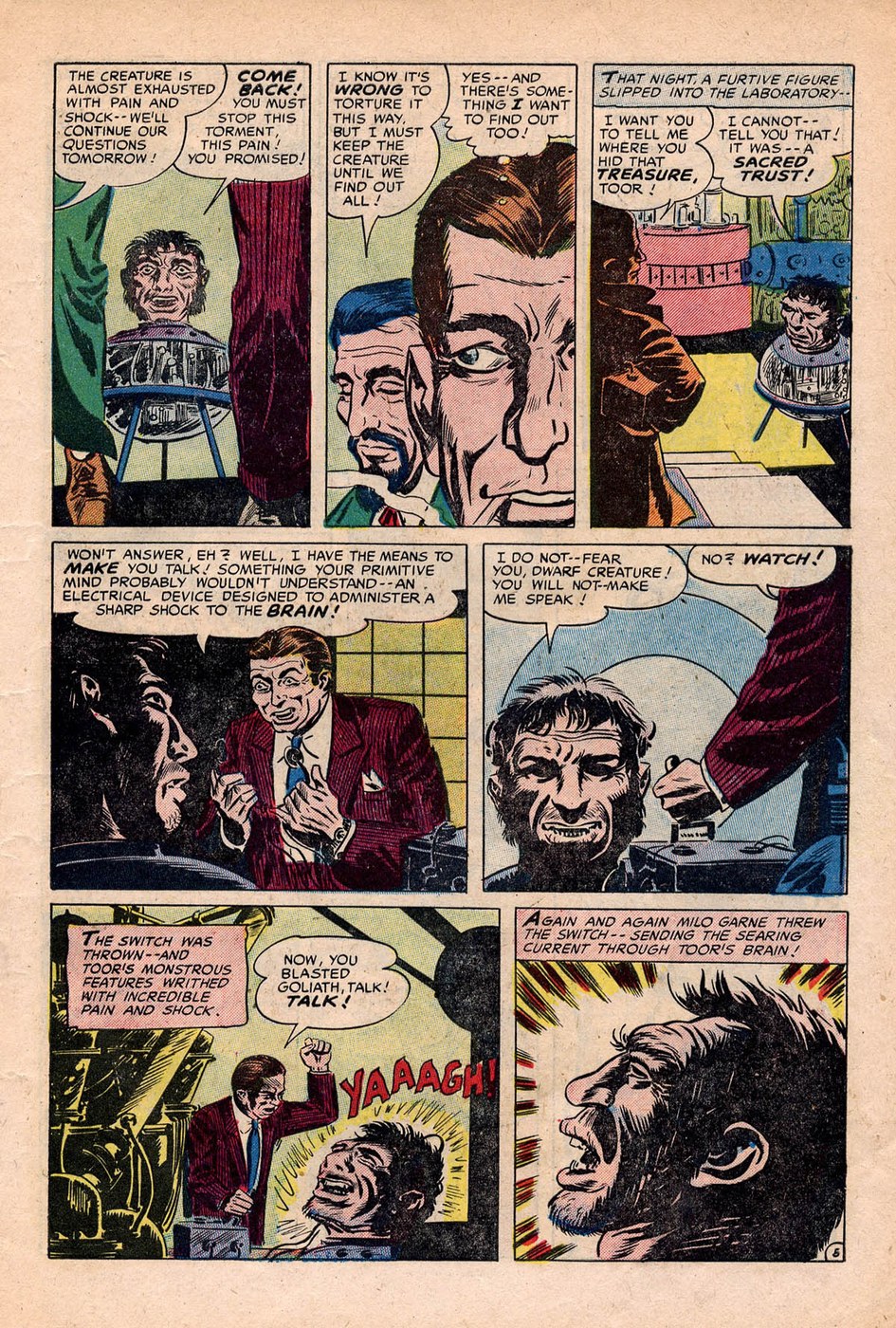}\caption{1953--54: \emph{Out of the Night} \#16, ACG, 1954, \texttt{80\_7}; 7 panels, 195 words.}\end{subfigure}
\hfill
\begin{subfigure}[t]{0.32\linewidth}\centering\includegraphics[width=1\linewidth,height=0.3\textheight,keepaspectratio]{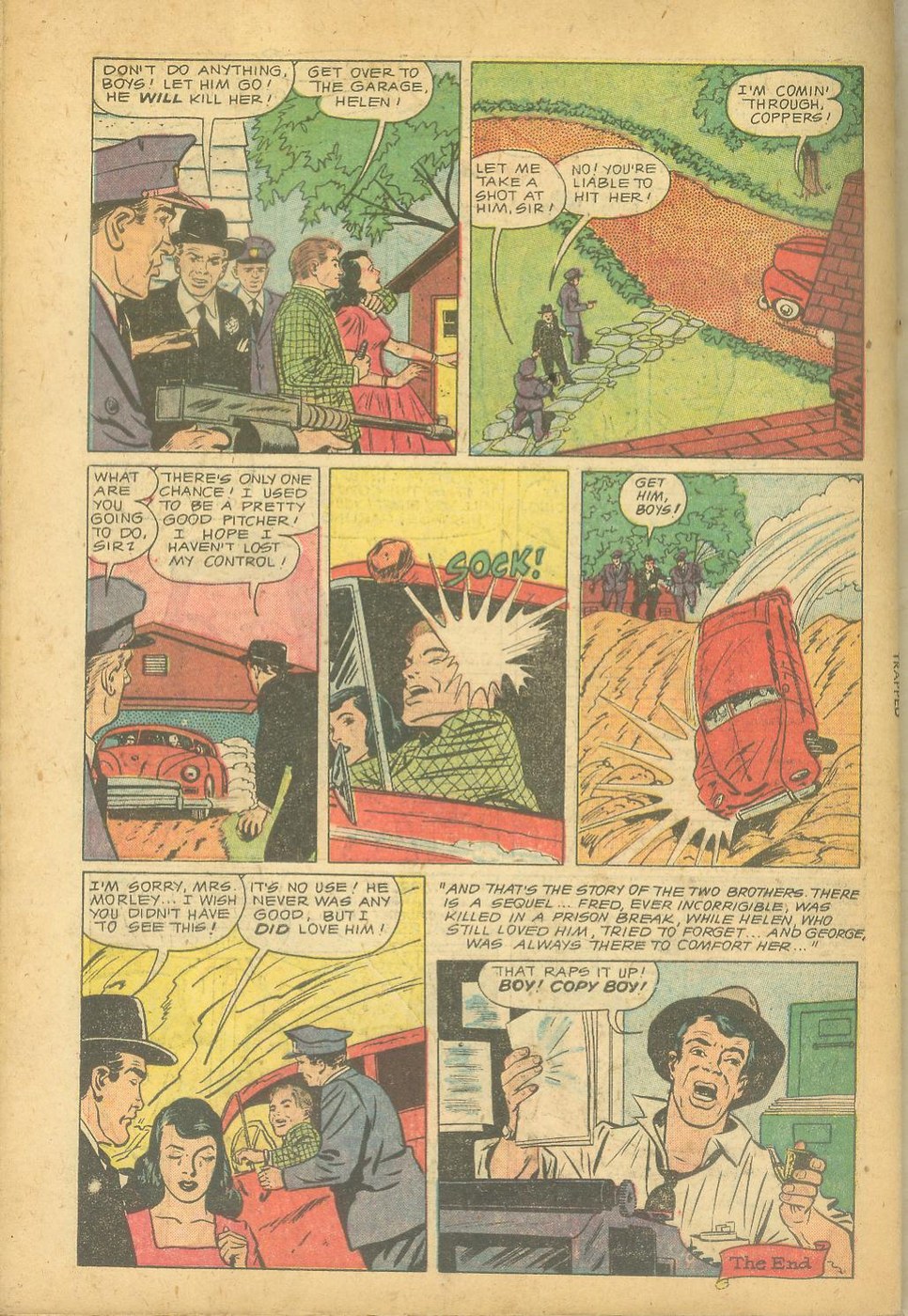}\caption{1953--54: \emph{Trapped!} \#1, Ace, 1954, \texttt{3085\_34}; 7 panels, 196 words.}\end{subfigure}
\par\medskip
\caption{Study~2: typical pages before, during and after the war. In each window, the two story pages of the speaker sample closest to the window's median panels and words per page (medians 1940--41: 8 panels and 142 words; 1944--45: 7 panels and 157 words; 1953--54: 7 panels and 195 words).}\label{fig:typical}\end{figure}
\begin{figure}[p]\centering
\begin{subfigure}[t]{0.45\linewidth}\centering\includegraphics[width=1\linewidth,height=0.3\textheight,keepaspectratio]{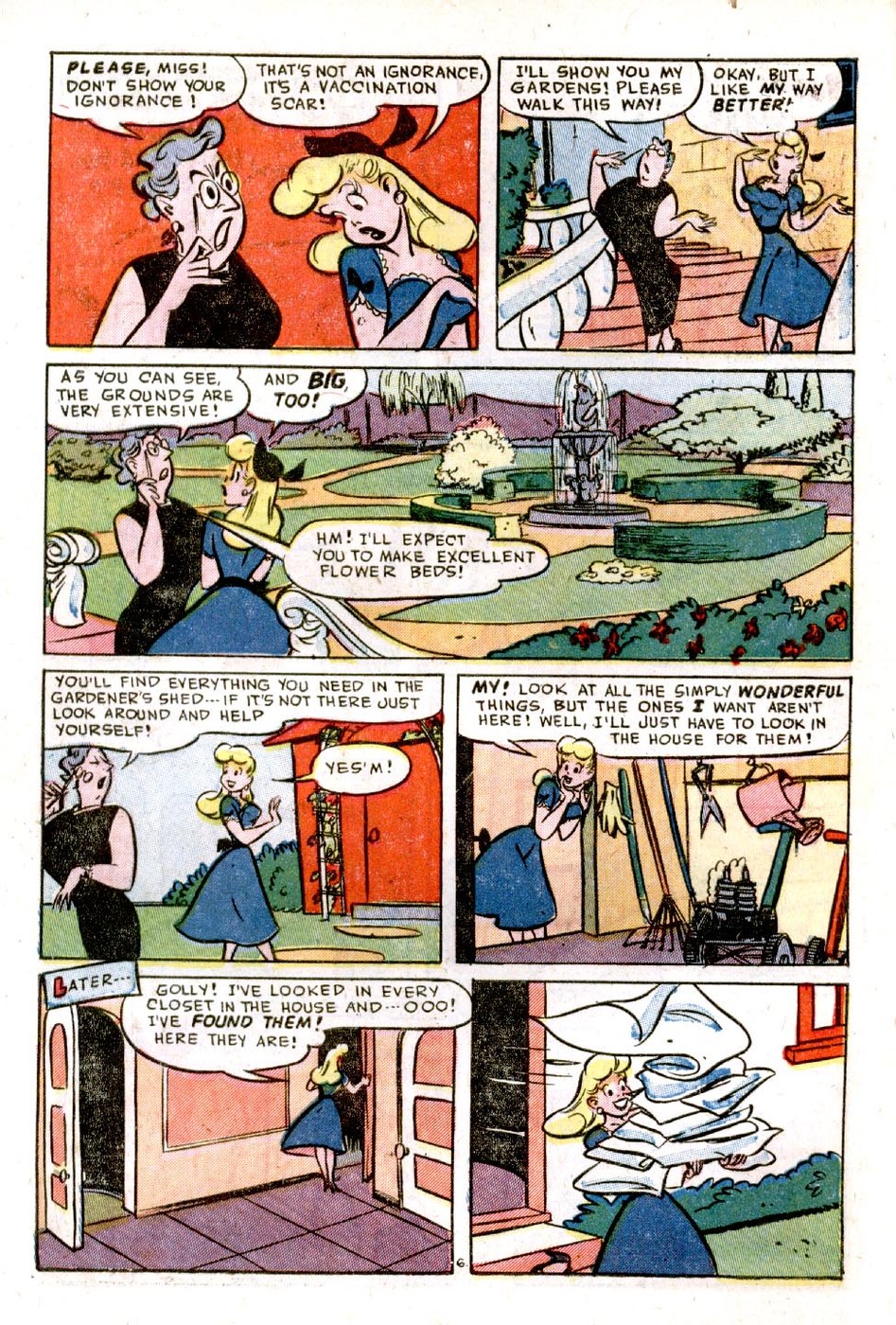}\caption{Title not matched, 1955, \texttt{990\_7}: the female-led page with the most balloons spoken by women. Coded speakers: 11 balloons by women, 0 by men; women speak 100\% of the words; drawn: 2 women, 0 men.}\end{subfigure}\hfill
\begin{subfigure}[t]{0.45\linewidth}\centering\includegraphics[width=1\linewidth,height=0.3\textheight,keepaspectratio]{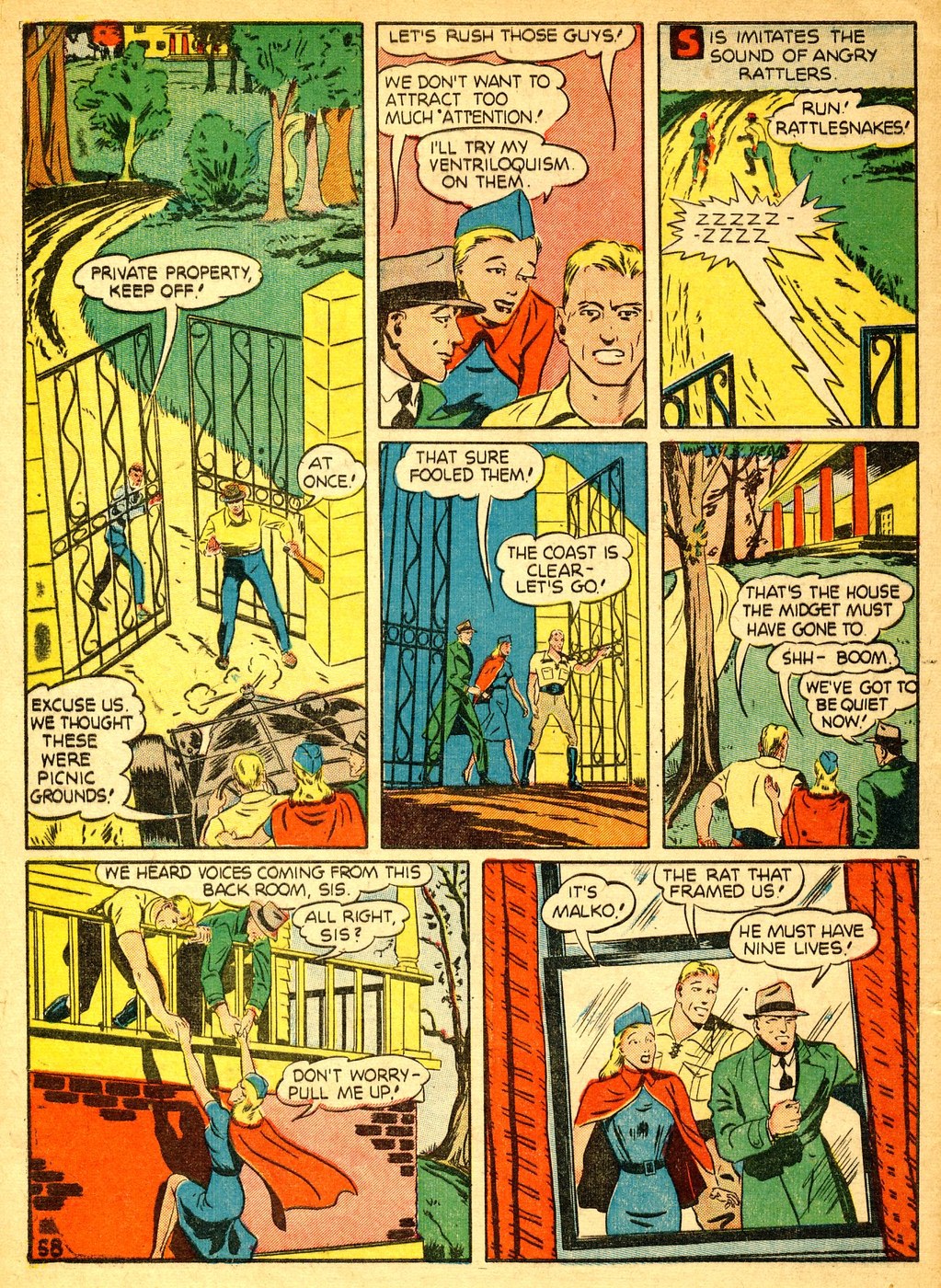}\caption{\emph{Power Comics} \#2, 1945, \texttt{2085\_58}: the page closest to the median women's word share among pages that draw a woman. Coded speakers: 4 balloons by women, 13 by men; women speak 23\% of the words; drawn: 1 woman, 4 men.}\end{subfigure}\hfill
\caption{Study~3: American story pages from the speaker sample, with the model's codes.}\label{fig:usvoice}\end{figure}
\begin{figure}[p]\centering
\begin{subfigure}[t]{0.48\linewidth}\centering\includegraphics[width=1\linewidth,height=0.36\textheight,keepaspectratio]{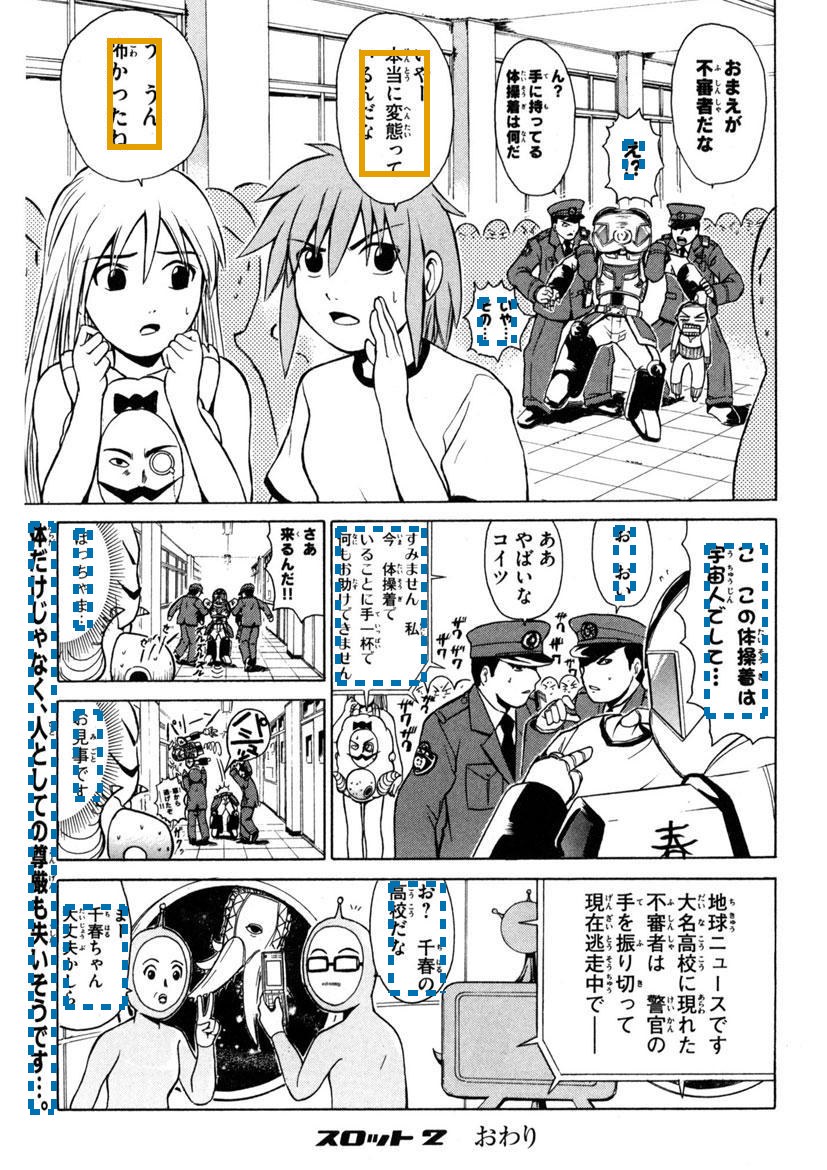}\caption{sh\=onen (boys): \begin{CJK}{UTF8}{ipxm}ランスロットフルスロットル\end{CJK}, 2000s; courtesy of \begin{CJK}{UTF8}{ipxm}ダイナミック 太郎\end{CJK}. Linked balloons: 2 by women, 10 by men.}\end{subfigure}
\hfill
\begin{subfigure}[t]{0.48\linewidth}\centering\includegraphics[width=1\linewidth,height=0.36\textheight,keepaspectratio]{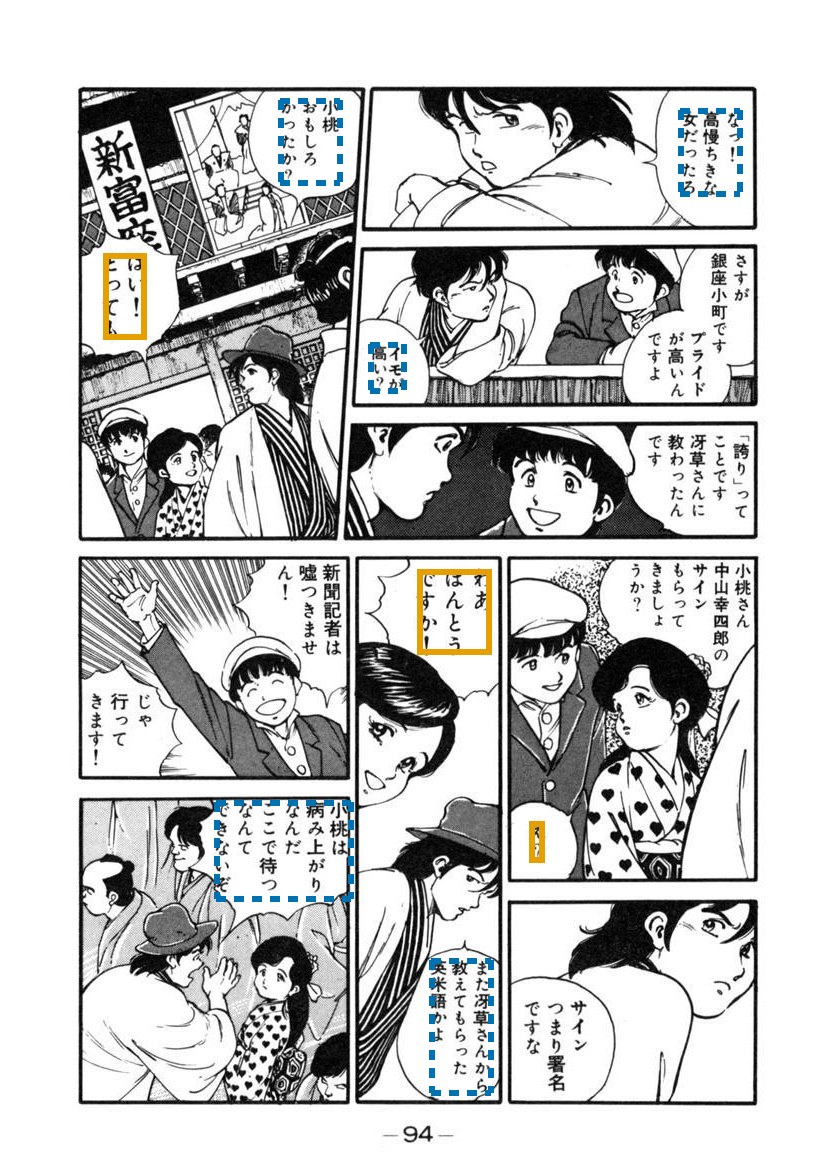}\caption{seinen (men): \begin{CJK}{UTF8}{ipxm}ヂャンギリぽんぽん\end{CJK}, 1980s; courtesy of \begin{CJK}{UTF8}{ipxm}木村 直巳\end{CJK}. Linked balloons: 3 by women, 5 by men.}\end{subfigure}
\par\medskip
\begin{subfigure}[t]{0.48\linewidth}\centering\includegraphics[width=1\linewidth,height=0.36\textheight,keepaspectratio]{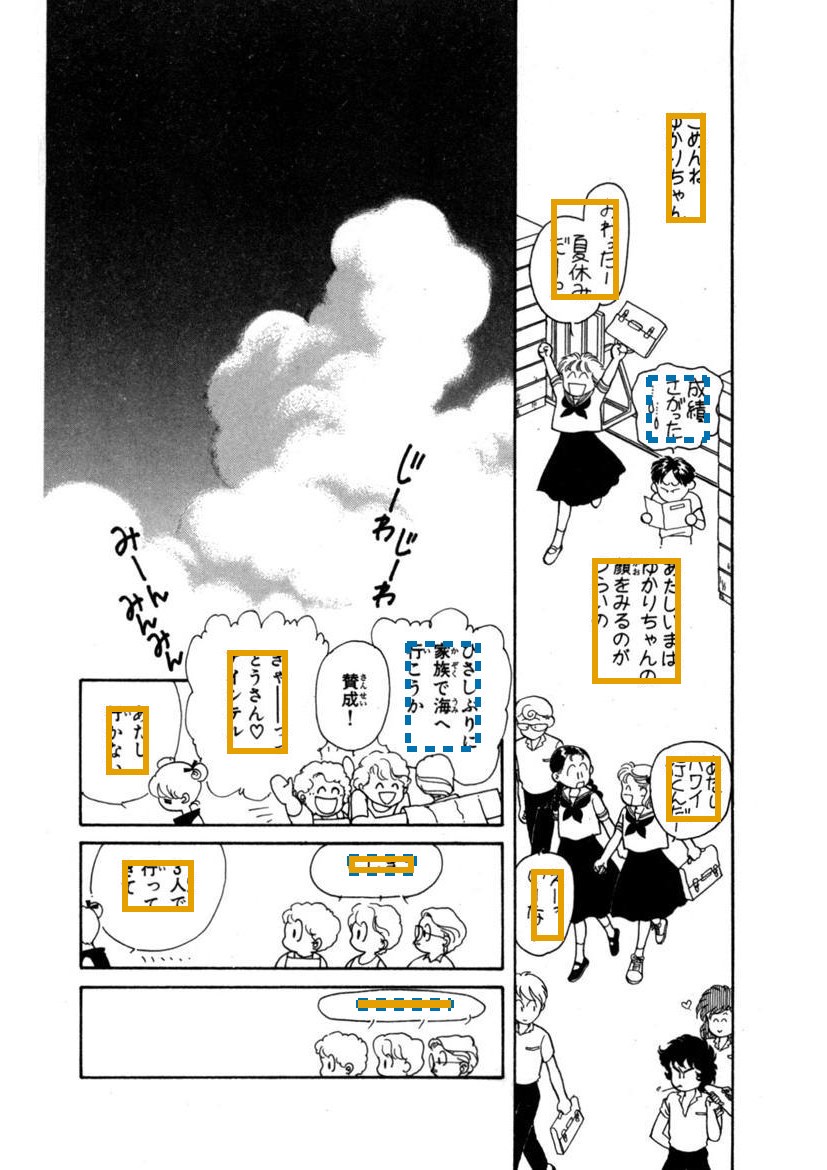}\caption{sh\=ojo (girls): \begin{CJK}{UTF8}{ipxm}その気でＡＢＣ\end{CJK}, 1980s; courtesy of \begin{CJK}{UTF8}{ipxm}たしろ 季霧\end{CJK}. Linked balloons: 10 by women, 4 by men.}\end{subfigure}
\hfill
\begin{subfigure}[t]{0.48\linewidth}\centering\includegraphics[width=1\linewidth,height=0.36\textheight,keepaspectratio]{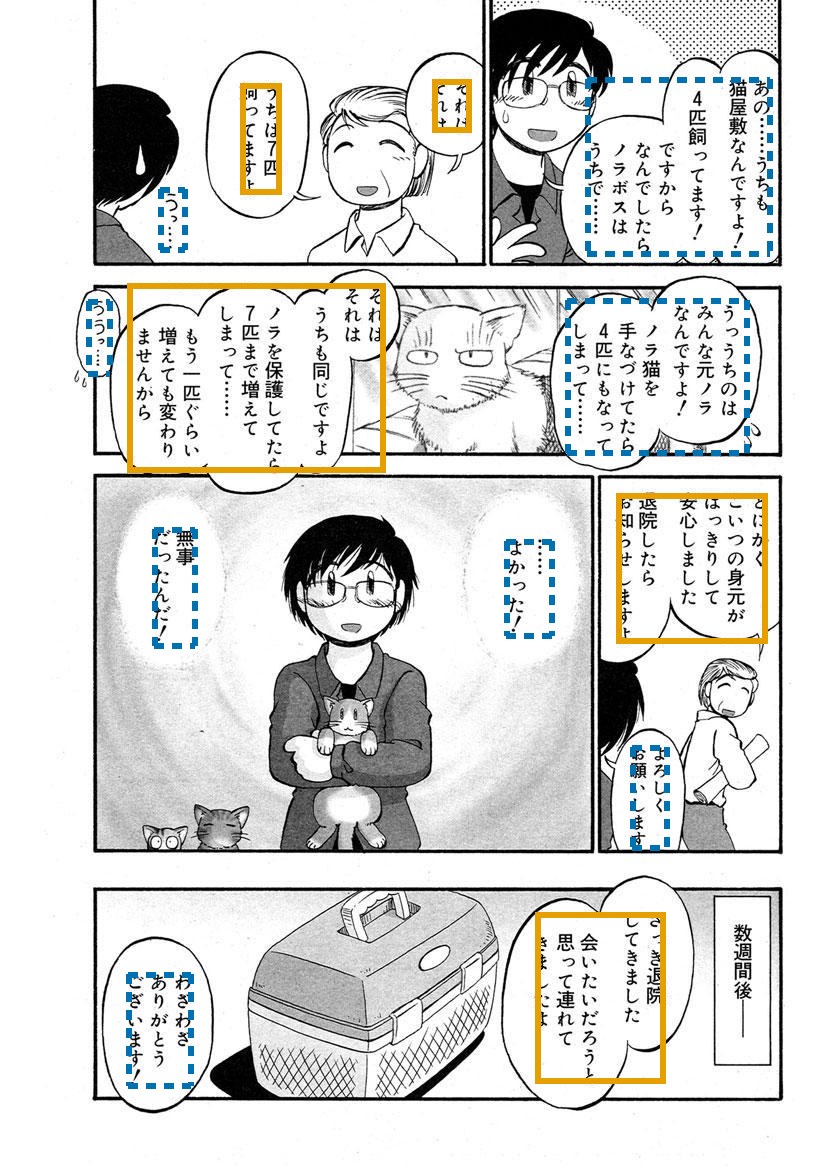}\caption{josei (women): \begin{CJK}{UTF8}{ipxm}うちの猫’ず日記\end{CJK}, 2000s; courtesy of \begin{CJK}{UTF8}{ipxm}がぁさん\end{CJK}. Linked balloons: 5 by women, 8 by men.}\end{subfigure}
\par\medskip
\caption{Study~3: one Manga109-s page per readership, with each balloon outlined by the coded gender of the speaker that Manga109Dialog annotators link to it (orange solid outline: woman; blue dashed outline: man). Each is the page with the most linked balloons, among the manga pages of its readership coded with the page-level procedure in Section~\ref{sec:grounding}, on which both women and men speak.}\label{fig:mangapages}\end{figure}
\clearpage

\end{appendices}

\bibliography{refs/comics_refs,refs/ferrara_refs}

\end{document}